\documentclass[letterpaper, 10 pt, conference]{ieeeconf}
\IEEEoverridecommandlockouts                              

\usepackage{cite}
\usepackage{amsmath}
\usepackage{graphicx}
\usepackage{float} 
\usepackage{subfigure}
\usepackage{overpic}
\usepackage{wrapfig}
\usepackage{balance} 
\usepackage{ulem}
\usepackage{color}
\usepackage[colorlinks,linkcolor=red]{hyperref} 
\usepackage{multirow}
\usepackage{mathtools}
\usepackage{booktabs} 
\usepackage{newtxmath}
\usepackage[linesnumbered, ruled]{algorithm2e}
\graphicspath{{figures/}}

\title{\LARGE \bf
Towards Large-Scale Incremental Dense Mapping using\\ Robot-centric Implicit Neural Representation
}

\author{Jianheng Liu and Haoyao Chen
\thanks{This work was supported in part by the National Natural Science Foundation of China (Grant No.U21A20119 and No.U1713206) and in part by the Shenzhen Science and Innovation Committee (Grant No.JCYJ20200109113412326 and No.JCYJ20210324120400003). (Corresponding author: Haoyao Chen.)}%
\thanks{J.H. Liu and H.Y. Chen* are with the School of Mechanical Engineering and Automation, Harbin Institute of Technology Shenzhen, P.R. China.
  {\tt\small liujianheng@stu.hit.edu.cn, hychen5@hit.edu.cn}.}%
}%

\begin{document}

\maketitle
\thispagestyle{empty}
\pagestyle{empty}

\begin{abstract}
Large-scale dense mapping is vital in robotics, digital twins, and virtual reality. Recently, implicit neural mapping has shown remarkable reconstruction quality.
However, incremental large-scale mapping with implicit neural representations remains problematic due to low efficiency, limited video memory, and the catastrophic forgetting phenomenon. To counter these challenges, we introduce the Robot-centric Implicit Mapping (RIM) technique for large-scale incremental dense mapping. This method employs a hybrid representation, encoding shapes with implicit features via a multi-resolution voxel map and decoding signed distance fields through a shallow MLP. We advocate for a robot-centric local map to boost model training efficiency and curb the catastrophic forgetting issue. A decoupled scalable global map is further developed to archive learned features for reuse and maintain constant video memory consumption. Validation experiments demonstrate our method's exceptional quality, efficiency, and adaptability across diverse scales and scenes over advanced dense mapping methods using range sensors. Our system's code will be accessible at \url{https://github.com/HITSZ-NRSL/RIM.git}.
\end{abstract}

\section{Introduction}

Large-scale dense mapping is integral to various autonomous robotics tasks, encompassing autonomous driving, surveying, and inspection. 
Storing geometric information in grid maps is common in robotics and 3D vision. 
Traditional mapping methods efficiently construct the map using diverse data structures like octree\cite{hornung2013octomap}, VDB\cite{museth2013vdb}, and hash map\cite{niessner2013real}. 
Recent works employ range sensors to realize efficient, high-quality reconstructions, like Voxblox\cite{oleynikova2017voxblox} and VDBFusion\cite{vizzo2022vdbfusion}. 
Although these methods guarantee good geometric accuracy, they struggle with high-granularity representation and scene speculation.
Conversely, the recent developments in implicit neural representations\cite{mescheder2019occupancy,mildenhall2021nerf} leverage multi-layer perceptrons (MLPs) to model 3D scene structures and advance in high-fidelity representation and scene speculation. 
More specific geometry attributes, like occupancy\cite{niemeyer2020differentiable} and SDF\cite{park2019deepsdf}, are introduced to delineate surfaces avoiding geometry ambiguities\cite{zhang2020nerf++}. 
iSDF\cite{ortiz2022isdf} train an MLP to correlate 3D coordinates with an approximate signed distance using a series of posed depth images. 
These pure MLP-based methods depict small scenes with tiny memory, but MLP's limited capacity constraints a detailed and broader scene depiction.

Explicit geometric structures\cite{liu2020neural,takikawa2021neural,muller2022instant} are integrated to embed learnable features in voxels to better sculpture scenes at the expense of memory consumption.
However, extending the implicit neural representations to large-scale scenes is challenging due to the limited video memory.
SHINE-Mapping\cite{zhong2022shine} harnesses the memory-efficient octree to reduce video memory consumption but still struggles to reconstruct vast scenes using mainstream devices with limited video memory.

Nonetheless, most implicit neural representations are trained in a batch manner, which is unsuitable for instant tasks. 
Incremental mapping supports streaming data processing, offering instantaneous feedback\cite{yan2023active} with greater flexibility than batch processing. 
However, incremental implicit neural mapping, as a continual learning problem, encounters the catastrophic forgetting phenomenon\cite{kaushik2021understanding}, where neural networks forget previously acquired knowledge during new learning phases. 
The idea of keyframes is introduced to hold a bundle adjustment for a consistent mapping\cite{sucar2021imap,zhu2022nice,ortiz2022isdf}. 
SHINE-Mapping\cite{zhong2022shine} employs a regularization-based approach to mitigate this challenge by confining the update direction of learnable features.
However, designing well-generalized keyframe selection strategies and selecting regularization parameters is tricky.

\begin{figure}[t]
    \centering
    \includegraphics[width=0.48\textwidth]{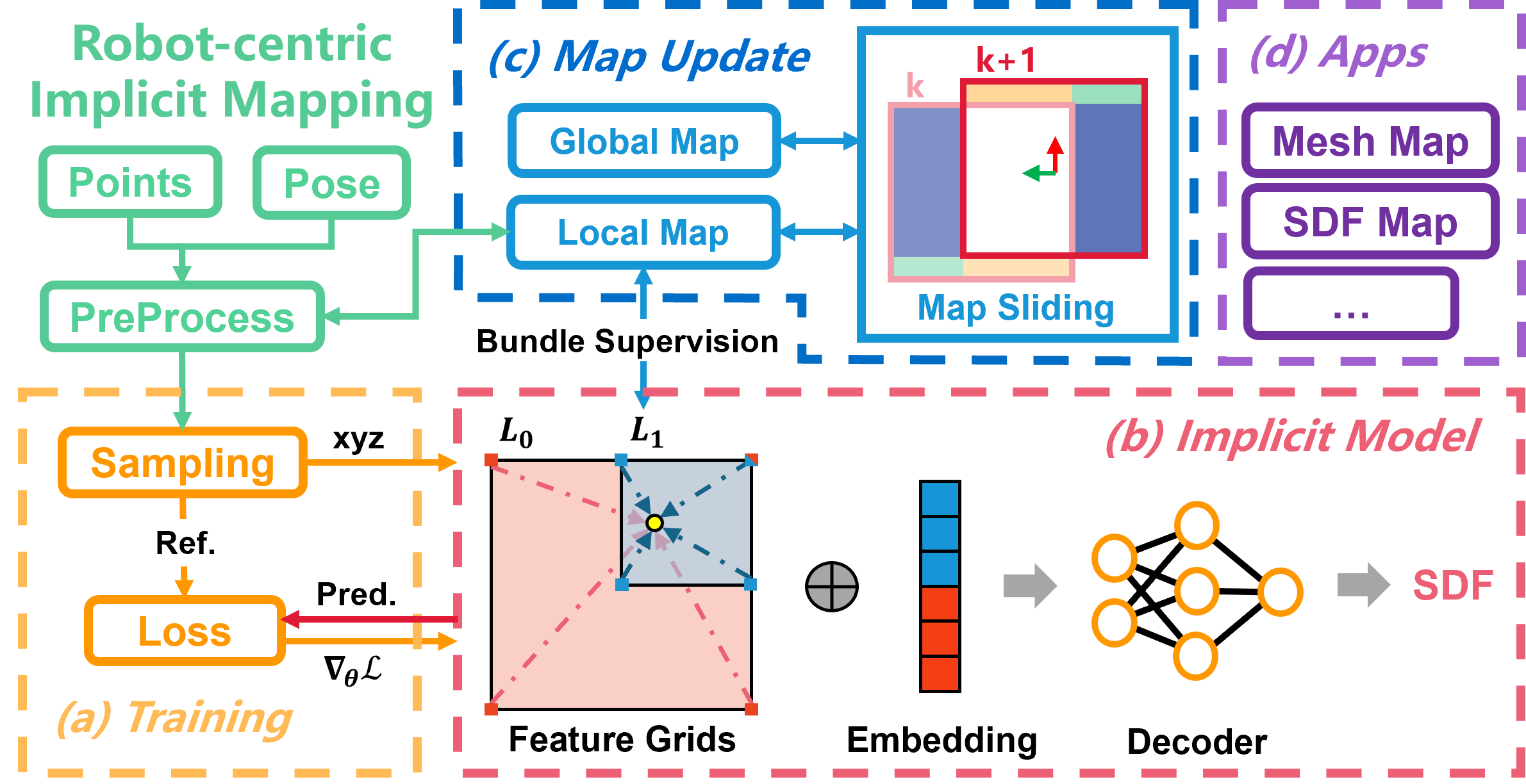}
    \caption{Pipeline of the robot-centric implicit mapping. RIM generates signed distance fields in real-time based on provided posed points.}
    \label{fig:pipeline}
    \vspace{-12pt}
\end{figure}

Reconstructing a large-scale dense map necessitates precise geometric information and demanding VRAM. 
Some implicit neural representations\cite{mildenhall2021nerf} can reconstruct high-fidelity synthetic views from RGB images alone, but they need much effort to present precise geometry. 
Range sensors effectively capture dense environmental geometrics and are extensively employed for pose estimation\cite{liu2022rgb}, detailed mapping\cite{newcombe2011kinectfusion} and motion planning\cite{wang2023active,wang2024whole}.
The incorporated depth information\cite{deng2022depth} also contributes to faster convergence and more refined geometric details in implicit neural representations.
Our method follows the dense mapping paradigm using range sensors.

To address the challenges posed by the MLP's limited capacity, escalating video memory consumption, catastrophic forgetting, and dynamic objects, we propose robot-centric implicit mapping (RIM) for large-scale incremental dense mapping. 
The overall pipeline is depicted in Fig.~\ref{fig:pipeline}. 
We embed learnable features in voxels and use a shallow MLP to infer signed distance field (SDF) for high-granularity representation, similar to voxel-based implicit representations\cite{takikawa2021neural}. 
We propose robot-centric mapping that trains the implicit model only on the local map with constant video memory consumption. 
It integrates numerous smaller local map training tasks into the whole global map training. 
To enable unbounded mapping, RIM crafts a global map using a hash map and sub-map set as a shelf to store and retrieve learned features from and for the local map. 
To mitigate the impact of catastrophic forgetting and dynamic objects, RIM leverages historical points within the local map to realize a multi-view bundle supervision training and outlier removal.

This paper's main contributions are threefold: 
\begin{enumerate}
    \item We propose a novel training framework for large-scale implicit neural representation using robot-centric mapping with constant video memory consumption.
    \item We exploit the traits of robot-centric mapping to conduct bundle supervision and outlier removal, achieving high-quality reconstruction and mitigating the catastrophic forgetting and dynamic object influence.
    \item We construct a flexible global map structure that allows dynamic unbounded map expansion without needing a preset map boundary.
\end{enumerate}
Validation experiments are performed to verify the superior quality, efficiency, and versatility of our method across varied scales and scenarios.

\section{Methodology}

Our work aims to reconstruct a high-granularity large-scale dense map from a continuous stream of poses and depth data. 
To accomplish this, we present our implicit neural representation for signed distance fields, which is trained in Euclidean space without the need for space compaction\cite{ortiz2022isdf}. 
Subsequently, we describe our proposed robot-centric mapping method for high-quality implicit model training.
The comprehensive frame integration process is shown in Alg.\ref{alg:overview}, with each step elaborated in subsequent sections.

The notations are defined as follows: $G, L$ denote the global and local coordinate systems; $\mathbf{x}, \mathbf{p}$ represent the robot's and point's positions, respectively.

\subsection{Implicit Neural Representation}\label{sec:implicit_repre}
\subsubsection{Implicit Model}\label{sec:model}

We adopt a hybrid representation that encodes shape with learnable features $\xi$ and decodes SDF values $\hat{d}$ using a shallow multi-layer perceptron, as illustrated in Fig.~\ref{fig:pipeline}(b).
The local map is composed of uniform feature voxels, where each voxel contains a learnable feature of dimensionality $D$.
We adopt a multi-resolution feature matrix with $L$ layers to integrate information at multiple granularities \cite{takikawa2021neural}. In this approach, the voxel size of the $l$-th layer map is determined by $2^{l-1}s$, where $s$ is the setting leaf voxel size. 
Given a point $\mathbf{p}^G_i \in \mathbb{R}^{3}$ that falls into the local map, we obtain its different-levels embedding feature ${\xi}_l(\mathbf{p}^G_i) \in \mathbb{R}^{D}$ by trilinear interpolation with its eight neighboring features. 
Different levels of implicit feature ${\xi}_l$ are concatenated into a new feature $\xi \in \mathbb{R}^{L\times D}$ integrated with multi-granularity information. 
An MLP-based decoder $f_\theta$ passes forward the concatenated feature and returns the predicted SDF value $\hat{d}_i$:
\begin{equation}
    \hat{d}_i = f_\theta \left( \bigoplus_{l=1}^L {\xi}_l \left( \mathbf{p}^G_i \right) \right),
\end{equation}
where $\bigoplus$ denotes the concatenation operation. We use $\hat{d}_i = f_\theta(\mathbf{p}^G_i)$ for brevity in the following sections.

\begin{algorithm}[t]
    \caption{Overall Frame Integration Process}
    \label{alg:overview}
    \KwIn{
        Current position $\mathbf{x}_k$, Input point cloud $\mathcal{P}_k$
    }
    \SetKwInput{Not}{Notation}
    \Not{Inside point cloud $\mathcal{P}_i$, Outside point cloud $\mathcal{P}_o$, Historical point cloud $\mathcal{P}_h$, Supervision set $\mathcal{S}$}{}
    \SetKwProg{Alg}{Algorithm}{}{}
    \Alg{}
    {
    $\mathtt{Slide}(\mathbf{x}_k);\ //$ \texttt{Sec.\ref{sec:structure}}\\
    $\mathcal{P}_h, \mathcal{P}_o = \mathtt{PreProcess}(\mathcal{P}_k);\ //$ \texttt{Sec.\ref{sec:forgetting_problem}}\\
    \ForEach{iteration}{
        $\mathcal{S} = \mathtt{Sample}(\mathcal{P}_h, \mathcal{P}_o);\ //$ \texttt{Sec.\ref{sec:sampling}}\\ \label{alg_sample}
        $\mathtt{Train}(\mathcal{S});\ //$ \texttt{Sec.\ref{sec:training}}\\ \label{alg_train}
        }
    }
    \textbf{End Algorithm}
    
    \SetKwProg{Fn}{Function}{}{}
    \Fn{$\mathtt{Slide(\mathbf{x},\mathcal{P})}$\label{alg_slide}}{
        $\mathtt{SaveOutsideBlockToGlobal}(\mathbf{x})$\;\label{alg_save}
        $\mathtt{UpdateLocalMapCenter}(\mathbf{x})$\;\label{alg_roll}
        $\mathtt{PadOutsideBlockFromGlobal}(\mathbf{x})$\;\label{alg_fetch}
    }
    \textbf{End Function}
    
    \Fn{$\mathtt{PreProcess(\mathcal{P})}$}{
        $\mathcal{P}_i,\mathcal{P}_o = \mathtt{SeperateInOutsidePoint}(\mathcal{P})$\;\label{alg_separate}
        $\mathtt{OutlierRemoval}(\mathcal{P}_h)$\;\label{alg_outlier_remove}
        $\mathcal{P}_h = \mathtt{UpdateHistoricalPoint}(\mathcal{P}_i)$\;\label{alg_hist}
        \Return{$\mathcal{P}_h, \mathcal{P}_o$}\;
    }
    \textbf{End Function}
    
    \Fn{$\mathtt{Sample(\mathcal{P}_h,\mathcal{P}_o)}$}{
        $\mathcal{S}_h = \mathtt{SampleHistoricalPoint}(\mathcal{P}_h)$\;\label{alg_sample_hist}
        $\mathcal{S}_o = \mathtt{SampleOutsidePoint}(\mathcal{P}_o)$\;\label{alg_sample_out}
        \Return{$\{\mathcal{S}_h, \mathcal{S}_o\}$}\;
    }
    \textbf{End Function}
\end{algorithm}

\subsubsection{Training}\label{sec:training}

\begin{figure}[t]
    \centering

  \setcounter{subfigure}{0}
    \subfigure[Spatial sampling]{
      \centering
      \includegraphics[width=0.225\textwidth]{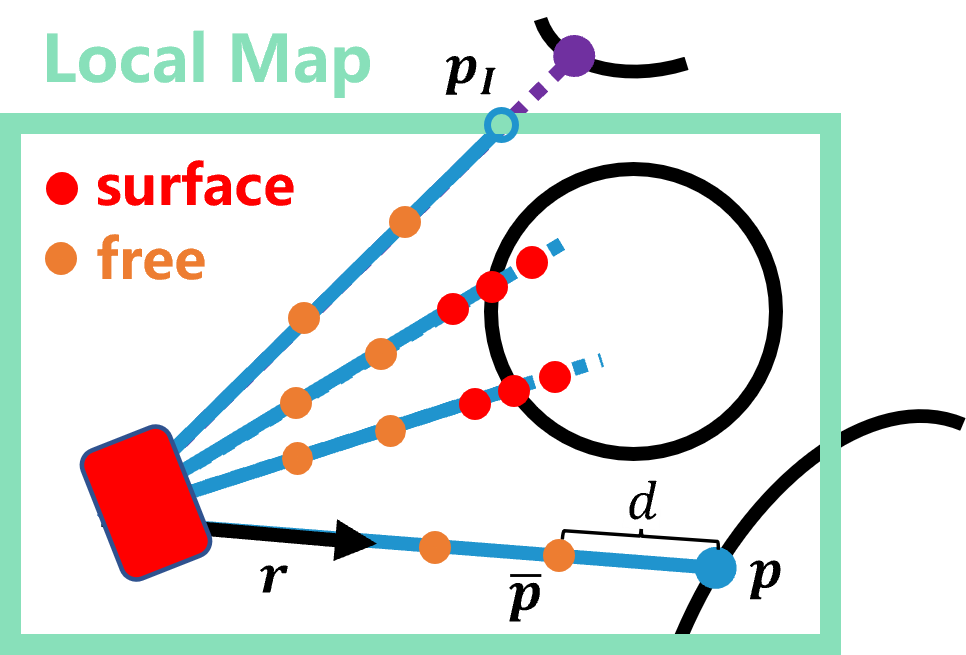}
  }
  \subfigure[Bundle supervision]{
      \centering
      \includegraphics[width=0.215\textwidth]{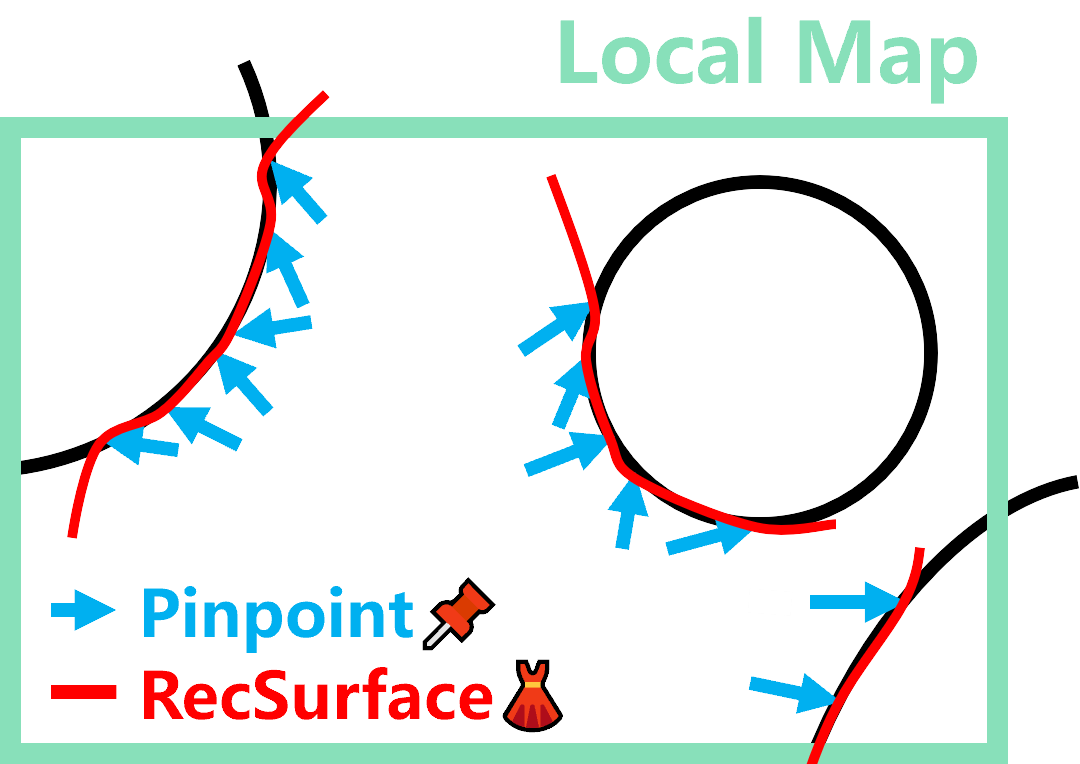}
  }
    \vspace{-4pt}

  \caption{The schematic diagram of spatial sampling. The high-quality input within the perceptual range carves the surface in detail, while the outside input distinguishes the free space. Moreover, the local map dynamically maintains historical points for bundle supervision.}
\label{fig:rim_sampling}
\vspace{-16pt}
\end{figure}

As illustrated in Fig.~\ref{fig:rim_sampling}(a), given any input point $\mathbf{p}$ and its ray direction $\mathbf{r}$, we sample points along its ray: $\bar{\mathbf{p}} = \mathbf{p} - {d} \mathbf{r}$, and form the supervision set $\mathcal{S} = \{\mathbf{p},d \}$. Each sample point's reference value $d$ defines the signed ray distance, which may differ from the true SDF value. Inspired by \cite{zhong2022shine}, we use the sigmoid function to map SDF to range $(0,1)$: $S(d) = 1/(1+e^{d/\sigma})$, where $\sigma$ is a hyperparameter to characterize the reconstruction smoothness and the sensor noise, and the $3\sigma$ distance can be considered as a soft truncated distance. The binary cross entropy is employed as the loss function as follows:
\begin{equation}
    \mathcal{L} = -\frac{1}{N}\sum_i^N o_i\log(\hat{o}_i) + (1-o_i)\log(1-\hat{o}_i),
\end{equation}
where $o_i = S({d}_i)$ is the reference value, $\hat{o}_i = S(f_\theta({\mathbf{p}}_i))$ is the implicit model's predicted value, and $N$ is the number of supervision points. 
The combination of the sigmoid function and the binary cross entropy adaptively gives higher weights to points near surfaces. 
The implicit model parameters are updated according to gradients $\nabla_\theta\mathcal{L}$ of the loss function (Alg.\ref{alg:overview}-Ln.\ref{alg_train}), as shown in Fig.~\ref{fig:pipeline}(a).


Random values are commonly used for parameter initialization to enhance learned features' generalization\cite{zhong2022shine}. 
In this paper, we initialize all voxel features with zeros and the decoder with random values, expecting features to have similar implicit representations of the same shapes so that the decoder can have a consistent input to ease its burden.

\subsection{Robot-centric Mapping}

\subsubsection{Map Structure}\label{sec:structure}

\begin{figure}[t]
    \centering
    \includegraphics[width=0.4\textwidth]{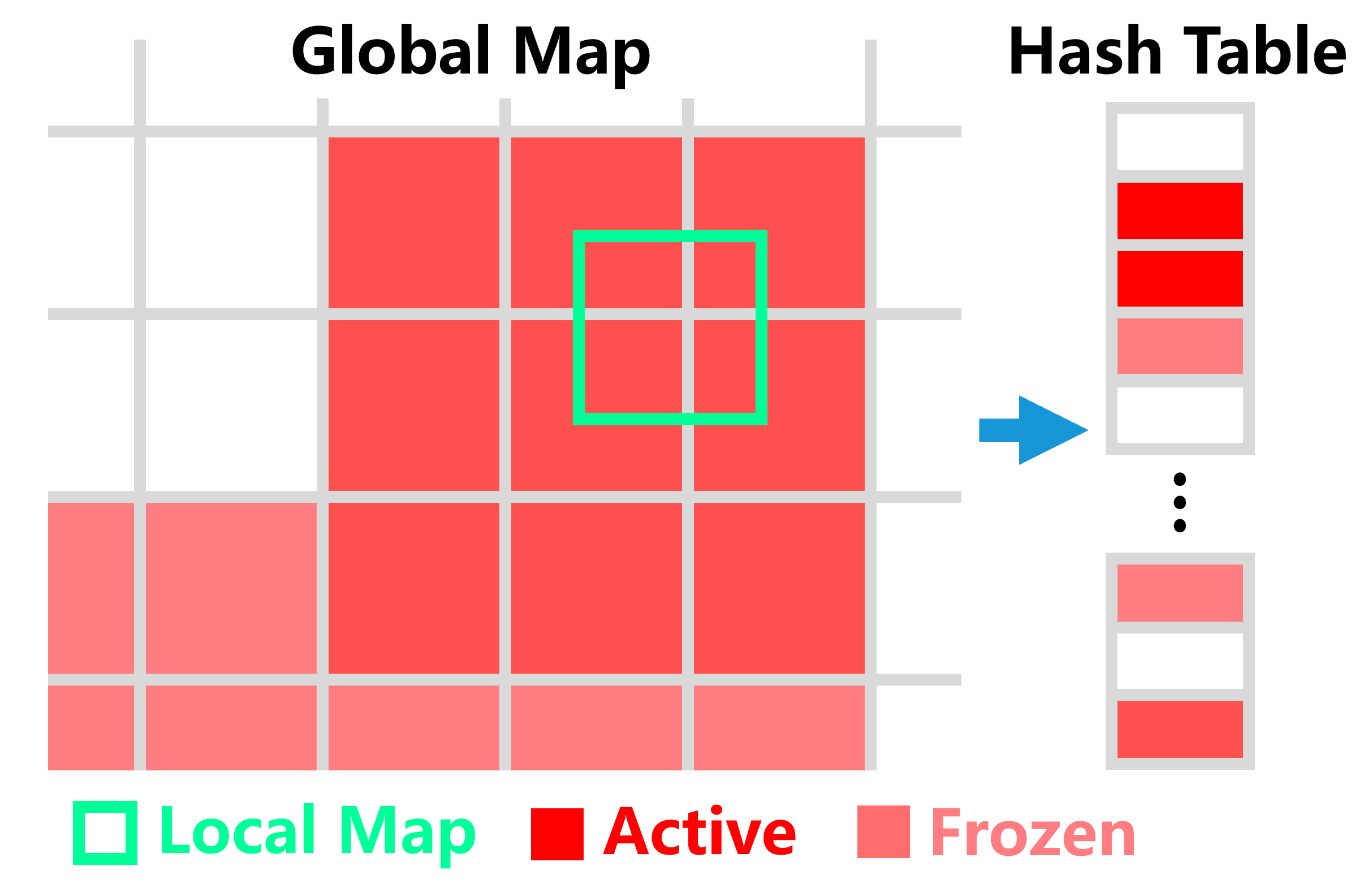}
    \caption{The schematic diagram of the implicit map structure and the sub-maps management. The local map slides in the global map. Each grid represents a sub-map, and those without color are non-allocated. A hash table arranges the allocated sub-maps. Red grids indicate active sub-maps stored in video memory, and pink grids indicate frozen sub-maps stored in system memory.}
    \label{fig:map_structure}
    \vspace{-16pt}
\end{figure}

In this paper, the implicit model is only trained on a robot-centric local map whose center aligns with the robot's position regardless of rotation. 
The local map's size is set according to the sensor's perceptual range and the robots' task space size. 
The local map slides the dense feature matrix to the new origin according to the robot's position without destroying or allocating memory (Alg.\ref{alg:overview}-Ln.\ref{alg_roll}), as shown in Fig.~\ref{fig:pipeline}(c) 's map sliding. 
A global map stores the learned features for reuse and is decoupled from the local map.
It consists of sub-maps and a shared decoder with the local map. 
A sub-map is a feature matrix the same size as the local map. 
As shown in Fig.~\ref{fig:map_structure}, every grid represents a sub-map, and the global map is covered with sub-maps without overlapping.

The local map is a cut of the global map, as shown in Fig.~\ref{fig:map_structure}, where each voxel in the local map has a unique mapping voxel in the global map:
\begin{equation}
    \mathbf{v}^G_i = \mathbf{v}^A_i - \mathbf{v}^A_L + \mathbf{v}^G_L,
\end{equation}
where $\mathbf{v}$ represents the voxel index and $A$ denotes local map matrix coordinate system.
With the local map moving to an unexpanded area, a new sub-map is created and stored in the global map using a hash table.
When the origin of the local map moves, blocks of voxels go out of the local map; and these blocks' features are cached to save into the corresponding sub-maps in the global map using a separate thread (Alg.\ref{alg:overview}-Ln.\ref{alg_save}). 
The newly expanded block is padded with fetched features from the global map (Alg.\ref{alg:overview}-Ln.\ref{alg_fetch}).

The robot-centric map structure decouples the training and storage maps using local and global maps.
It allows us to reconstruct a large-scale scene with constant video memory. 
As shown in Fig.~\ref{fig:map_structure}, we use a freeze-activate mechanism to transfer sub-maps between video and system memory. We keep the local map's nearby sub-maps activated for fast feature access. When a sub-map is far from the current pose, we freeze it by moving it to the system memory. When the robot revisits a historically frozen sub-map, we reactivate it by moving it back to video memory. 

\subsubsection{Bundle Supervision for Incremental Mapping}\label{sec:forgetting_problem}

Incremental neural mapping as a continual learning method suffers from catastrophic forgetting problems\cite{kaushik2021understanding}.
Voxel-based implicit representations mitigate this problem by decoupling part of implicit features using the explicit geometric representation; however, due to the property of shared features between voxels, the historical features still might be degenerated by adjacent new inputs\cite{zhong2022shine}.

\begin{table*}[!t]
    \centering
    \caption{Quantitative reconstruction results of 8 scenes on the Replica dataset. Our approach yields the best reconstruction accuracy (C-L1) and completeness (F-Score) over other methods.}
    \label{tab:replica_quan}
    \resizebox{1.0\textwidth}{!}{
    \begin{tabular}{cccccccccc}
        \toprule
        \textbf{Metrics} & \textbf{Methods} & \textbf{Office-0} & \textbf{Office-1} & \textbf{Office-2} & \textbf{Office-3} & \textbf{Office-4} & \textbf{Room-0} & \textbf{Room-1} & \textbf{Room-2} \\
        \midrule
        
        \multirow{5}*{Acc.[cm]$\downarrow$}  
        & {Voxblox} 
        & \underline{1.08} & \underline{0.85} & 1.03 & \underline{1.06} & 0.93 & \underline{0.91} & \underline{0.80} & 1.04\\ 
        & {VDBFusion} 
        & \textbf{0.56} & \textbf{0.53} & \textbf{0.56} & \textbf{0.58} & \textbf{0.58} & \textbf{0.59} & \textbf{0.54} & \textbf{0.56} \\ 
        & {iSDF} 
        & 2.02 & 2.07 & 2.07 & 2.01 & 1.65 & 1.86 & 1.31 & 2.20 \\ 
        & {SHINE} 
        & 2.23 & 1.67 & 2.31 & 2.15 & 1.48 & 2.01 & 2.45 & 1.99 \\  
        & \textbf{Ours} 
        & {1.12} & 0.93 & \underline{0.99} & \underline{1.06} & \underline{0.83} & 1.01 & 0.85 & \underline{0.94} \\ 
        \midrule

        \multirow{5}*{Comp.[cm]$\downarrow$}  
        & {Voxblox} 
        & 10.35 & 9.70 & 5.19 & 3.16 & 3.43 & 3.09 & 2.92 & 4.22 \\ 
        & {VDBFusion} 
        & {9.75} & 9.07 & 4.71 & {2.97} & {3.18} & {2.96} & {2.81} & {3.79} \\ 
        & {iSDF} 
        & 12.90 & 11.89 & 14.16 & 3.74 & 5.55 & 3.09 & 2.72 & 9.02 \\ 
        & {SHINE} 
        & \textbf{8.57} & \underline{7.68} & \underline{4.23} & \underline{2.89} & \underline{3.01} & \underline{2.58} & \underline{2.25} & \underline{3.44} \\ 
        & \textbf{Ours} 
        & \underline{8.61} & \textbf{7.49} & \textbf{3.78} & \textbf{2.45} & \textbf{2.65} & \textbf{2.26} & \textbf{2.14} & \textbf{3.09} \\ 
        
        \midrule

        \multirow{5}*{C-L1[cm]$\downarrow$}  
        & {Voxblox} 
        & 5.71 & 5.27 & 3.11 & 2.11 & 2.18 & 2.00 & 1.86 & 2.63 \\
        & {VDBFusion} 
        & \underline{5.15} & 4.80 & \underline{2.64} & \underline{1.77} & \underline{1.88} & \underline{1.77} & \underline{1.68} & \underline{2.17} \\
        & {iSDF} 
        & 7.46 & 6.98 & 8.12 & 2.87 & 3.60 & 2.48 & 2.01 & 5.61 \\
        & {SHINE} 
        & 5.40 & \underline{4.68} & 3.27 & 2.52 & 2.24 & 2.30 & 2.35 & 2.72 \\
        & \textbf{Ours} 
        & \textbf{4.86} & \textbf{4.21} & \textbf{2.39} & \textbf{1.75} & \textbf{1.74} & \textbf{1.63} & \textbf{1.46} & \textbf{2.02} \\
        \midrule

        \multirow{5}*{F-Score[\%]$\uparrow$}  
        & {Voxblox} 
        & 91.15 & 89.35 & 92.14 & 94.26 & 94.10 & \underline{95.43} & 95.58 & 93.81 \\
        & {VDBFusion} 
        & \underline{91.80} & \underline{90.33} & \underline{93.21} & \underline{94.64} & \underline{94.62} & {95.35} & 95.40 & \underline{94.68} \\
        & {iSDF}
        & 88.67 & 88.02 & 86.42 & 92.65 & 91.36 & 95.03 & \underline{96.16} & 89.18 \\
        & {SHINE} 
        & 88.70 & 89.86 & 91.59 & 93.54 & 93.54 & 94.47 & 90.85 & 92.90 \\
        & \textbf{Ours} 
        & \textbf{92.49} & \textbf{91.81} & \textbf{94.44} & \textbf{95.71} & \textbf{95.67} & \textbf{96.65} & \textbf{96.90} & \textbf{95.76} \\

        \bottomrule
    \end{tabular}
    }
    \vspace{-12pt}
\end{table*}

To address this problem, we separate the input point cloud into the inside point cloud $\mathcal{P}_i$ and outside point cloud $\mathcal{P}_o$ according to the local map's scope (Alg.\ref{alg:overview}-Ln.\ref{alg_separate}).
The inside point cloud $\mathcal{P}_i$ is accumulated into historical point cloud $\mathcal{P}_h$ to conduct bundle supervision in training (Sec.~\ref{sec:training}). 
We discard historical points that slide out of the local map and randomly drop exceeding points to keep a constant point number for fixed memory usage (Alg.\ref{alg:overview}-Ln.\ref{alg_hist}).
We consider points as pins and the reconstruction surfaces as clothes, as shown in Fig.~\ref{fig:rim_sampling}(b). 
Each pin fixes the local reconstructed cloth from deviation.

The proposed bundle supervision is more like point cloud registration than keyframe-based methods, maintaining point rays instead of view attitudes to constrain learned features with multi-views.
It does not need delicate keyframe selection strategies or keyframe maintenance in keyframe-based methods\cite{ortiz2022isdf} or training efficiency sacrifice in regularization-based methods\cite{zhong2022shine}. 
However, the decoder's parameters also update continuously during the training, which can cause inconsistency between learned features over time. 
Therefore, we fix the decoder parameters after learning a certain number of frames.

\subsubsection{Saptial Sampling and Outlier Removal}\label{sec:sampling}

Supervision points are sampled from the surface and the free (non-occupied) space along the ray to train a consistent implicit map. 
As shown in Fig.~\ref{fig:rim_sampling}(a), we sample historical inside points and current input outside points for each training iteration (Alg.\ref{alg:overview}-Ln.\ref{alg_sample}).
For inside point rays in $\mathcal{P}_i$, we use depth-guided sampling: $N_s$ surface points' sampling follows the normal distribution $\mathcal{N}(d,\sigma^2)$ and $N_f$ free points are stratified sampled between surfaces and the sensor (Alg.\ref{alg:overview}-Ln.\ref{alg_sample_hist}). 
For outside point rays in $\mathcal{P}_o$, we calculate their intersections $\mathbf{p}_I$ (Fig.~\ref{fig:rim_sampling}) with the local map and stratified sample points between intersections and the sensor (Alg.\ref{alg:overview}-Ln.\ref{alg_sample_out}).

The free and outside point supervision helps RIM to mitigate the influence of dynamic objects and outliers by further applying outlier removal to the historical point cloud (Alg.\ref{alg:overview}-Ln.\ref{alg_outlier_remove}).
The proposed outlier removal periodically infers the SDF values of historical points and drop points whose SDF values are $\epsilon$ away from 0, where the zero level set of the SDF defines the fitting surface.
In the following training, the previously learned outlier features will be forgotten with the free and outside point supervision, as shown in Fig.~\ref{fig:lio_rim_qual}.


\section{Experiments}

This section conducts qualitative and quantitative experiments on public datasets to demonstrate the novelty and effectiveness of the proposed methods. We compare our method with state-of-the-art dense mapping methods: Voxblox\cite{oleynikova2017voxblox} and VDBFusion\cite{vizzo2022vdbfusion} based on traditional mapping methods, and iSDF\cite{ortiz2022isdf} and SHINE-Mapping\cite{zhong2022shine} based on implicit neural representations; and all methods reconstruct scenes incrementally with the same voxel size. 
We recover the implicit model to triangular mesh using marching cubes\cite{lorensen1987marching}. 
Since the proposed method cannot map surfaces outside the local map, we use a cropped ground truth map for a fair comparison.
The widely used reconstruction metrics are used for quantitative evaluation\cite{mescheder2019occupancy,zhong2022shine}: they are mapping accuracy (Acc., cm), completeness (Comp., cm), chamfer-L1 distance (C-L1, cm), and F-score ($<$10cm, \%).


The RIM experimental parameters, namely $L$, $D$, $\epsilon$, $N$, $N_{s}$, and $N_{f}$, are set to 3, 8, 5cm, 2048, 3, and 3, respectively.
The MLP used in our experiments has only a single hidden layer with dimension 32 with ReLU activations in the intermediate layer. 
We form our implicit model using Libtorch and train it using the Adam optimizer. 
All experiments are conducted on a platform equipped with an Intel i5-12600KF CPU with 32GB system memory and an NVIDIA RTX 3070 Ti GPU with 8GB video memory.

\begin{figure*}[!t]
    \subfigure{
    \centering
    \includegraphics[width=0.162\textwidth]{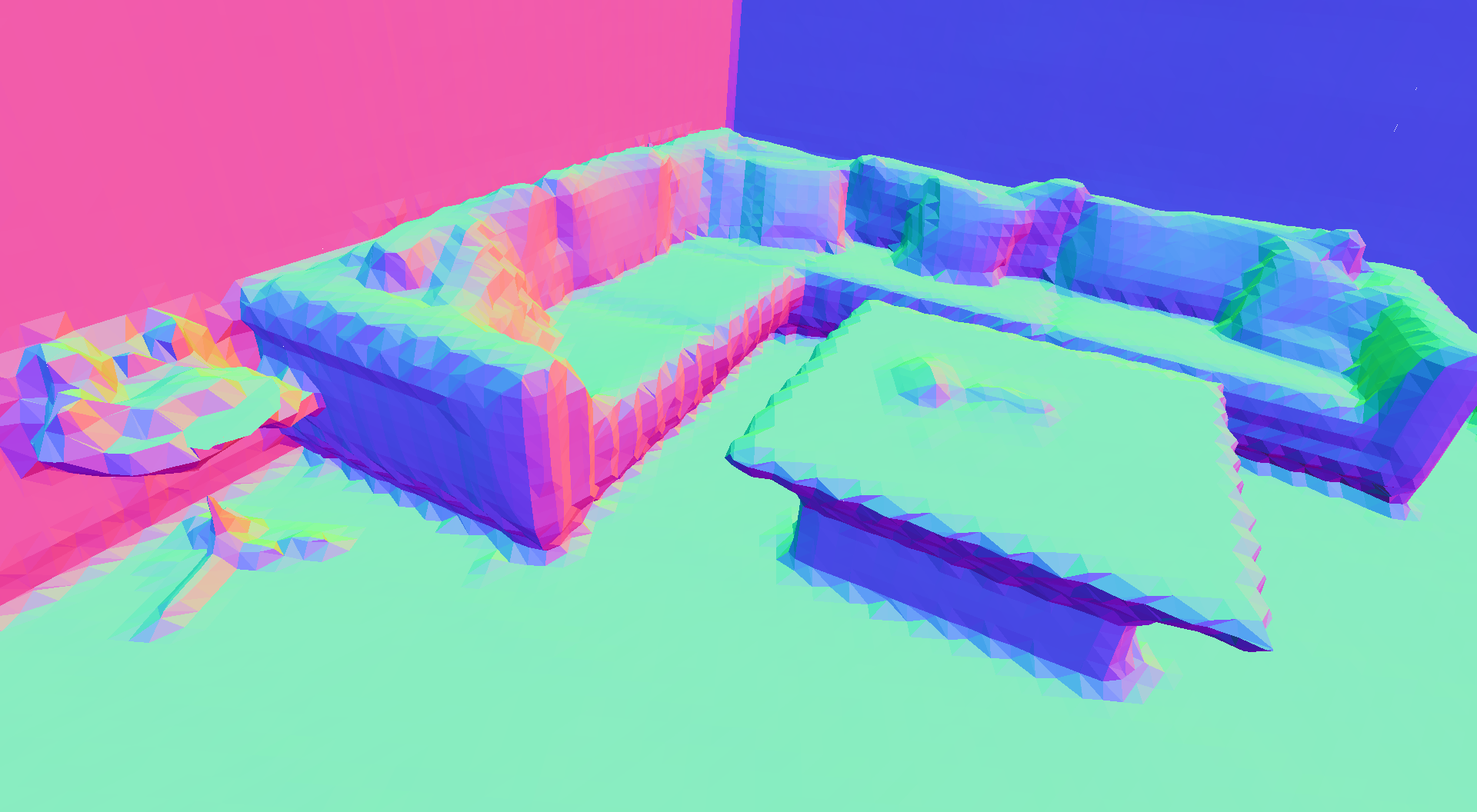}
    }
    \hspace{-11pt}
    \subfigure{
    \centering
    \includegraphics[width=0.162\textwidth]{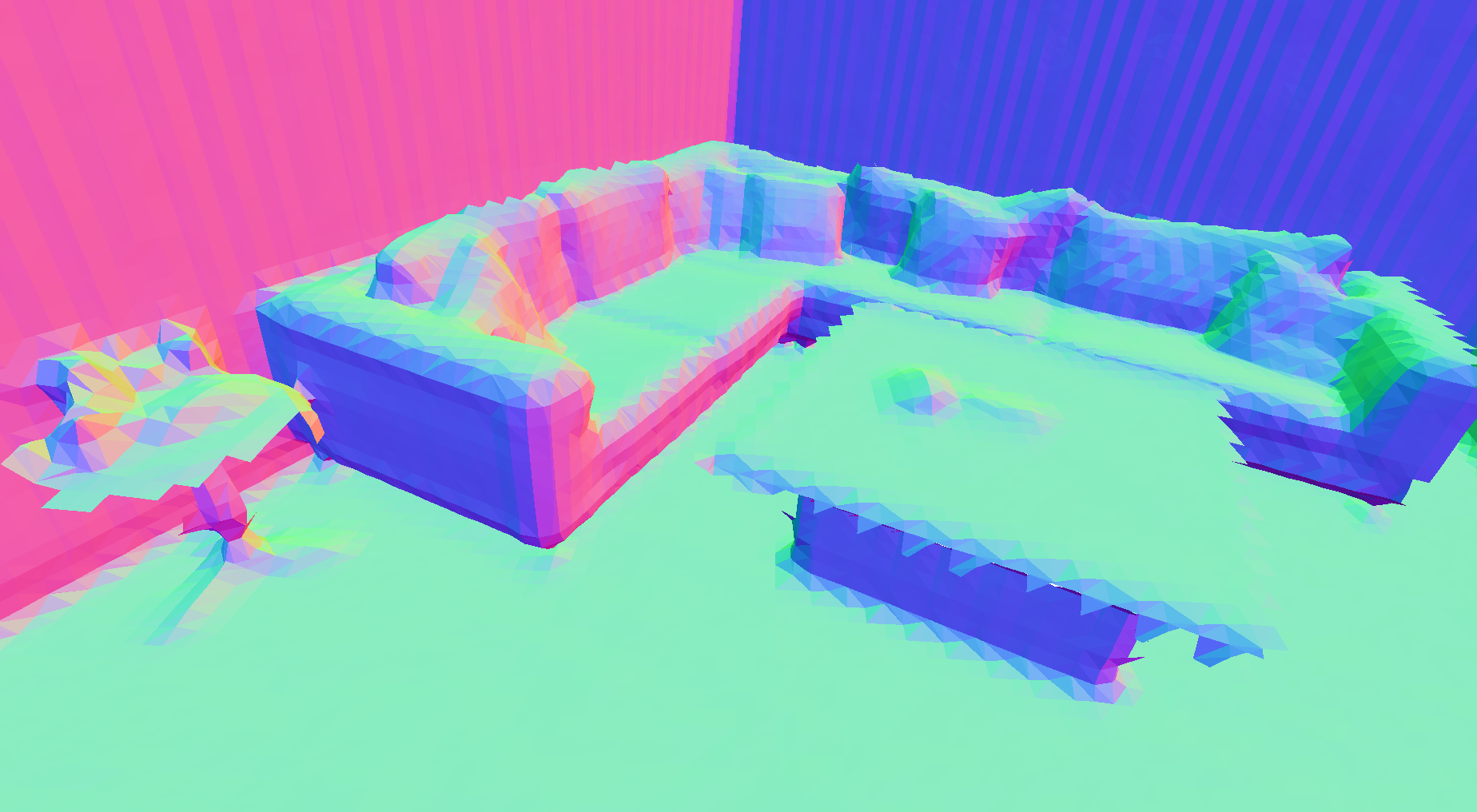}
    }
    \hspace{-11pt}
    \subfigure{
    \centering
    \includegraphics[width=0.162\textwidth]{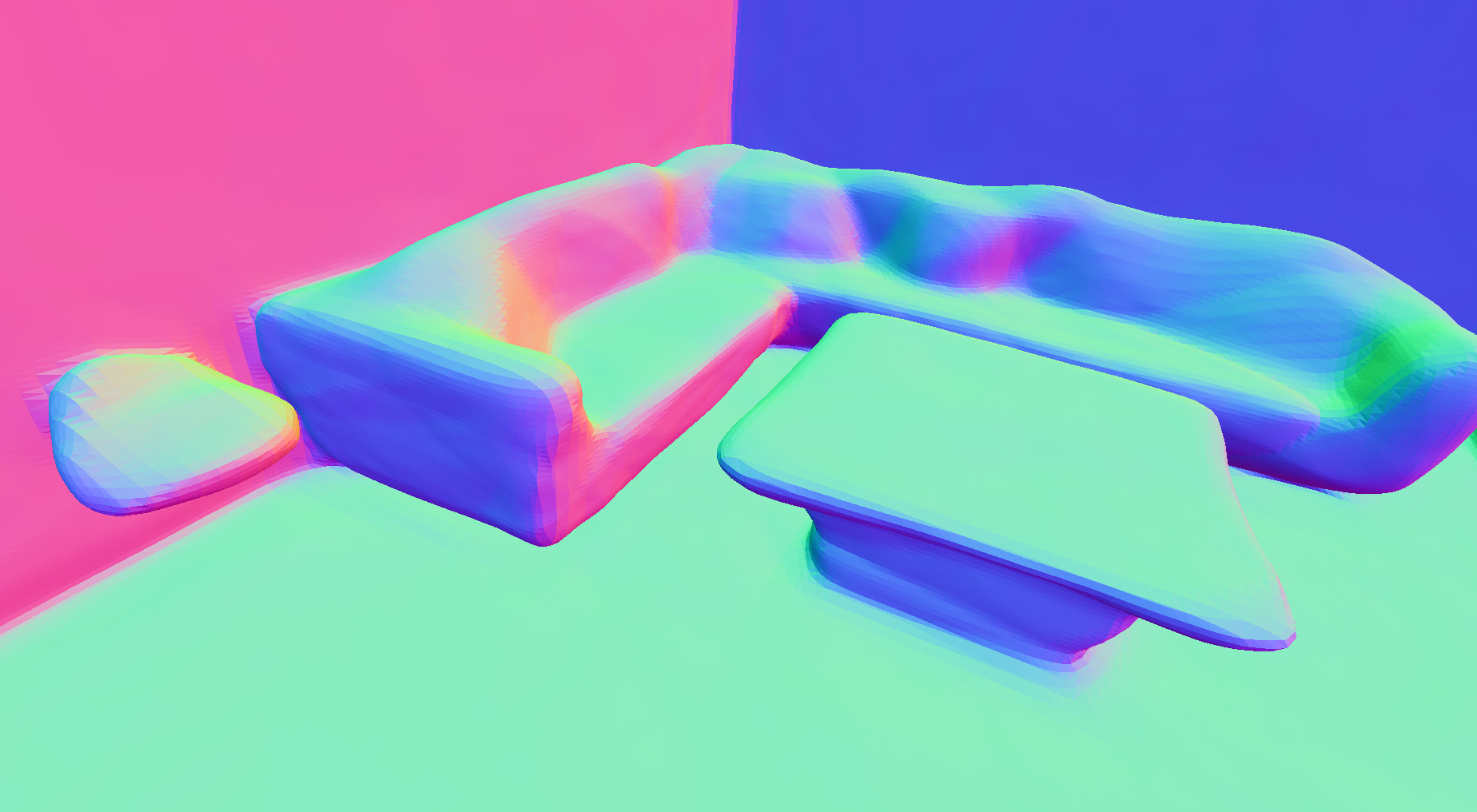}
    }
    \hspace{-11pt}
    \subfigure{
    \centering
    \includegraphics[width=0.162\textwidth]{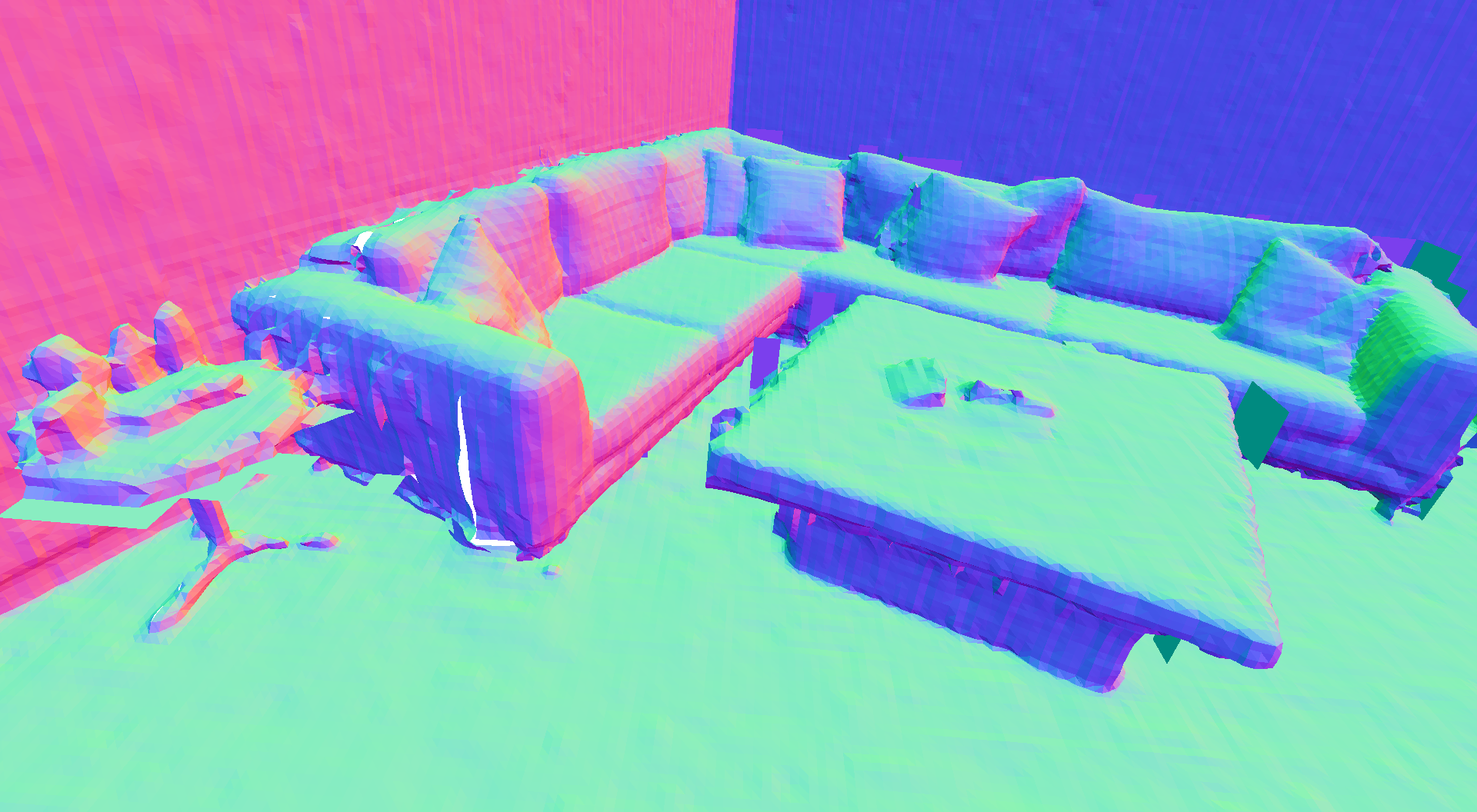}
    }
    \hspace{-11pt}
    \subfigure{
    \centering
    \includegraphics[width=0.162\textwidth]{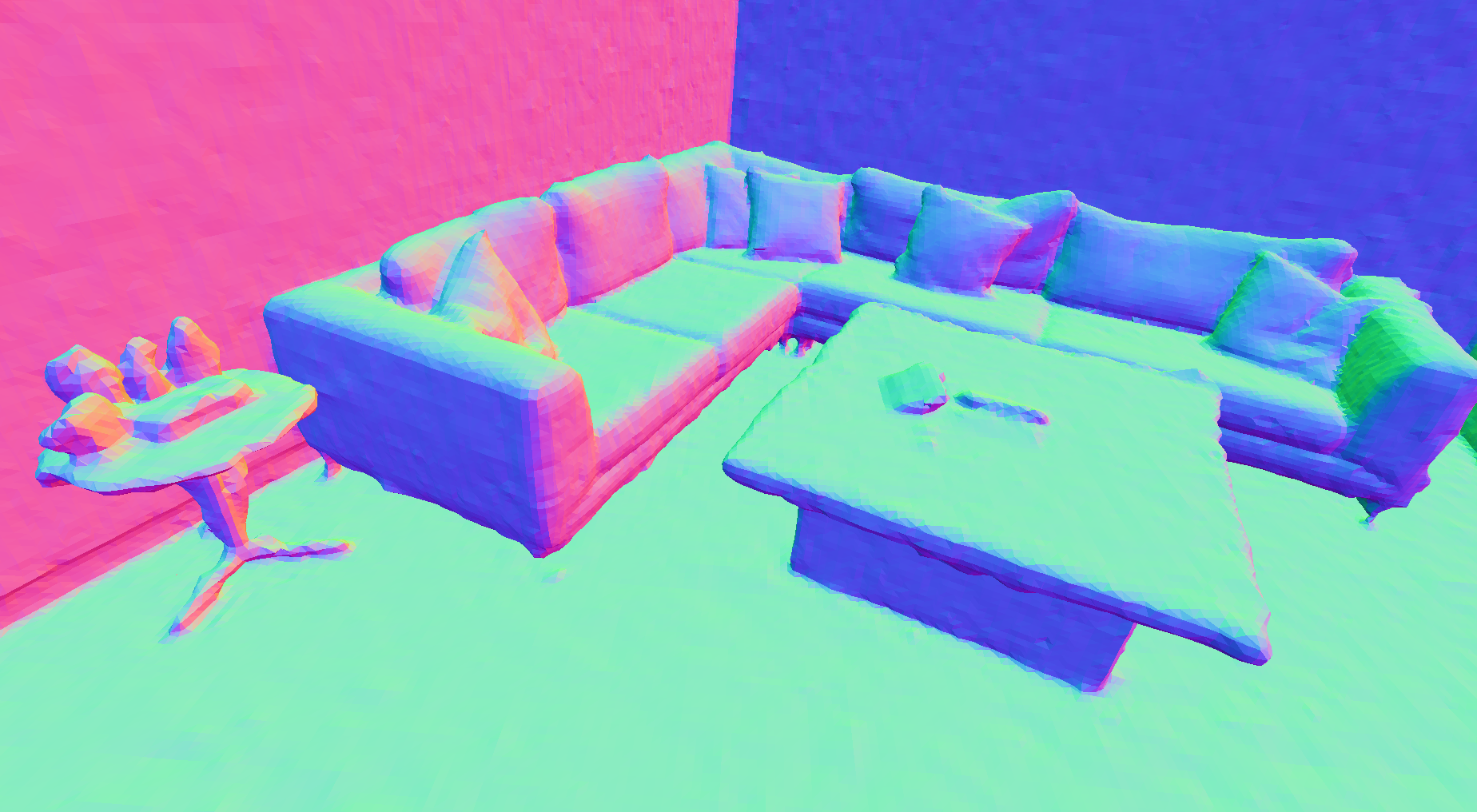}
    }
    \hspace{-11pt}
    \subfigure{
    \centering
    \includegraphics[width=0.162\textwidth]{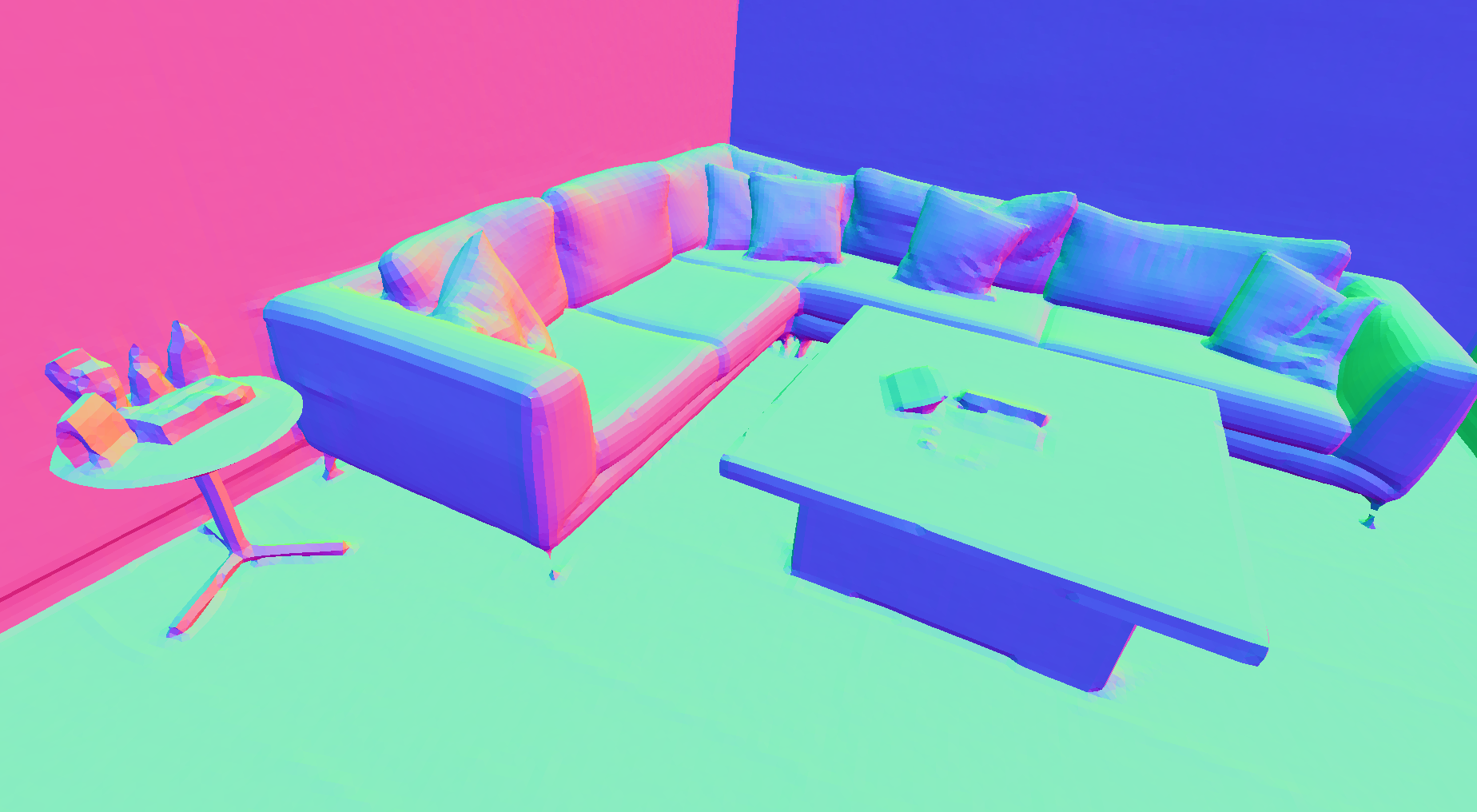}
    }
    
    \vspace{-8pt}
    
    \setcounter{subfigure}{0}
    \subfigure[Voxblox]{
    \centering
    \includegraphics[width=0.162\textwidth]{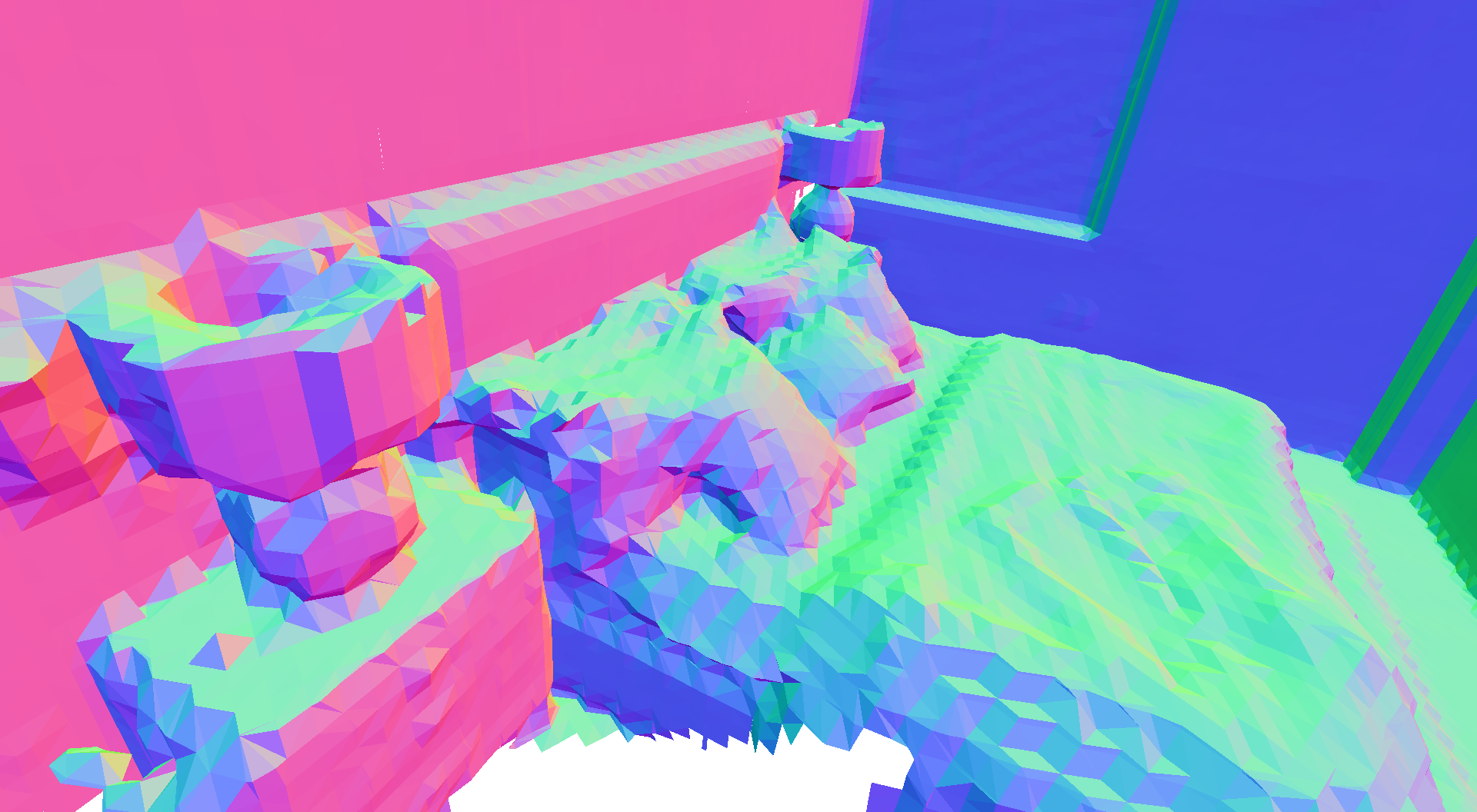}
    }
    \hspace{-11pt}
    \subfigure[VDB-Fusion]{
    \centering
    \includegraphics[width=0.162\textwidth]{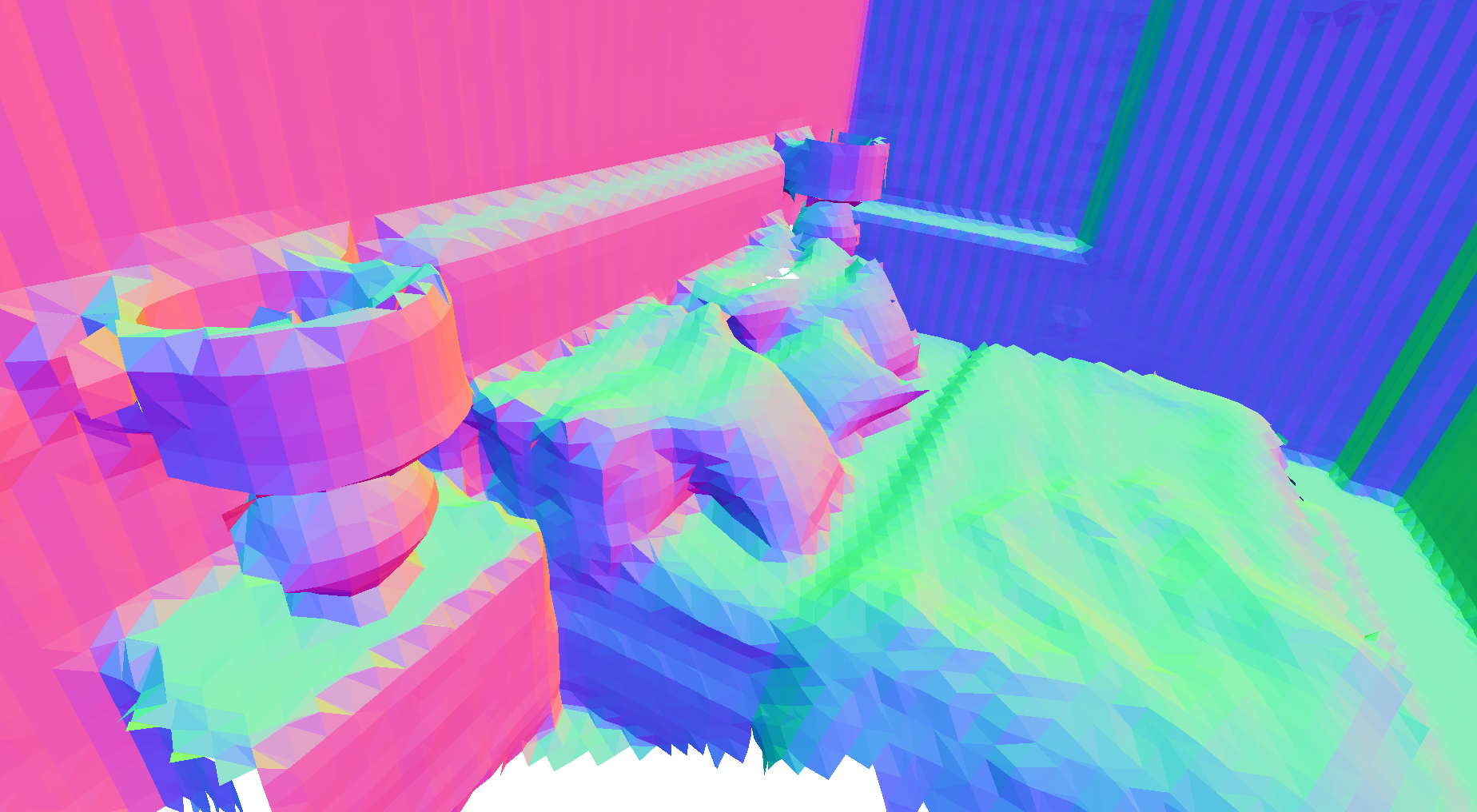}
    }
    \hspace{-11pt}
    \subfigure[iSDF]{
    \centering
    \includegraphics[width=0.162\textwidth]{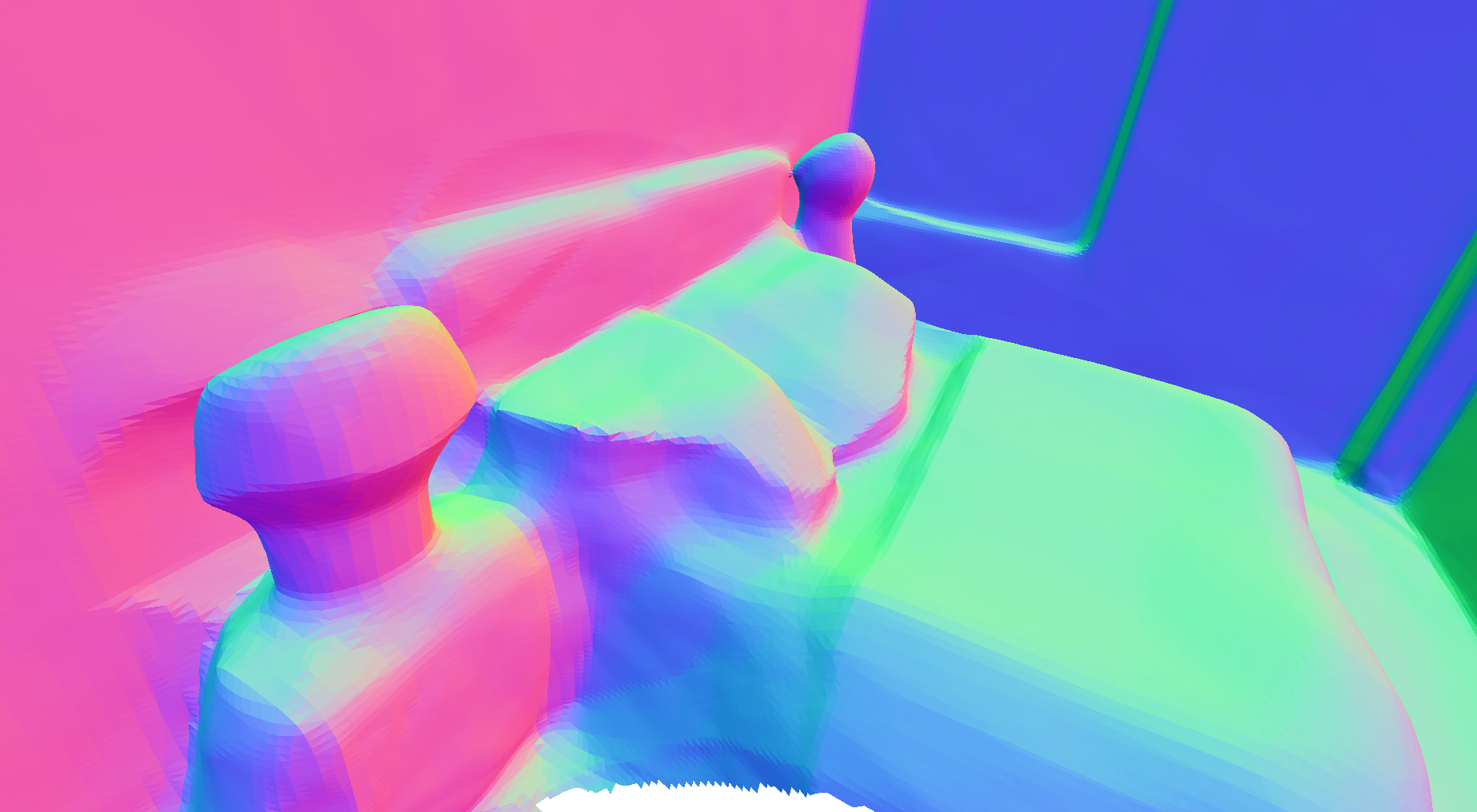}
    }
    \hspace{-11pt}
    \subfigure[SHINE-Mapping]{
    \centering
    \includegraphics[width=0.162\textwidth]{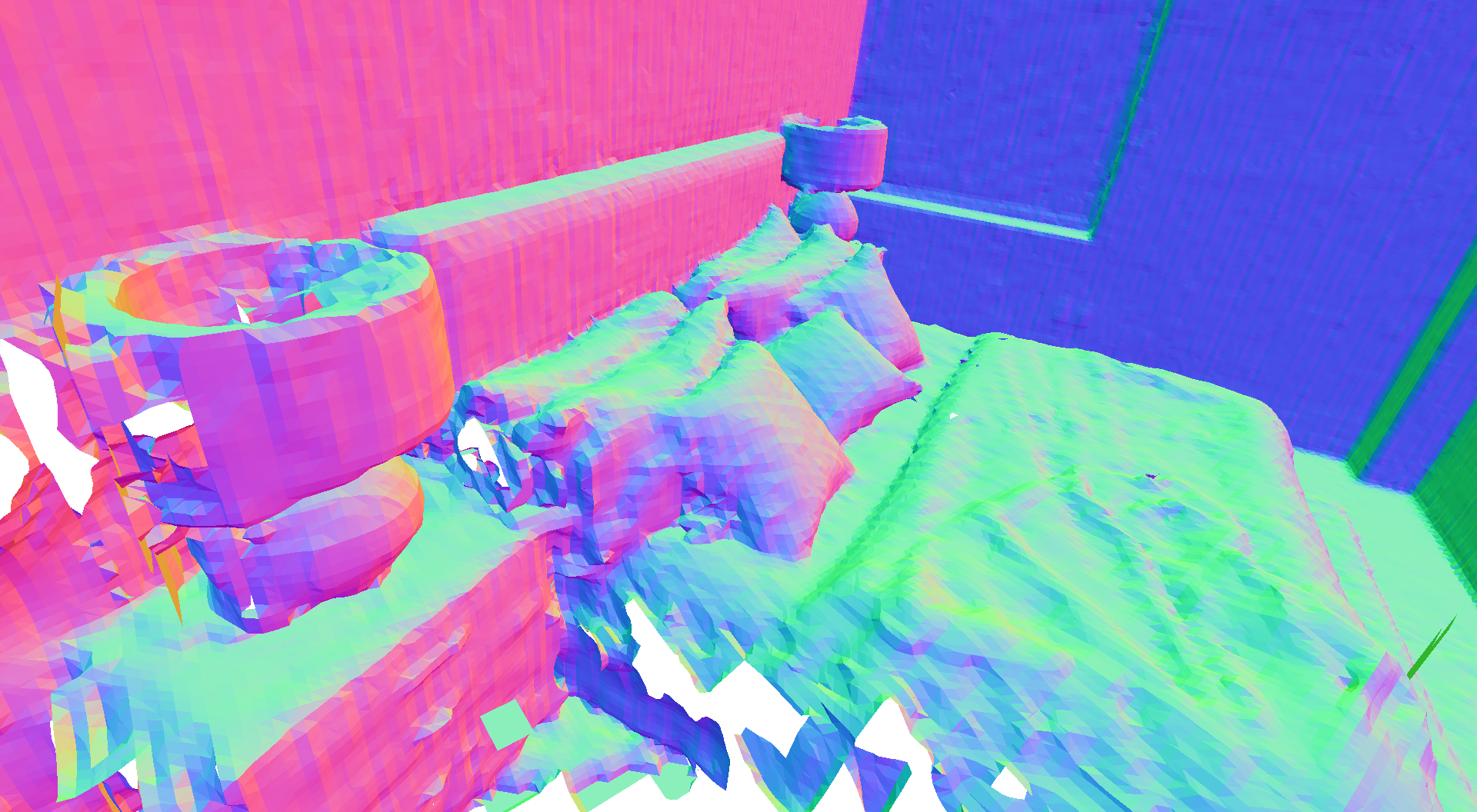}
    }
    \hspace{-11pt}
    \subfigure[Ours]{
        \centering
        \includegraphics[width=0.162\textwidth]{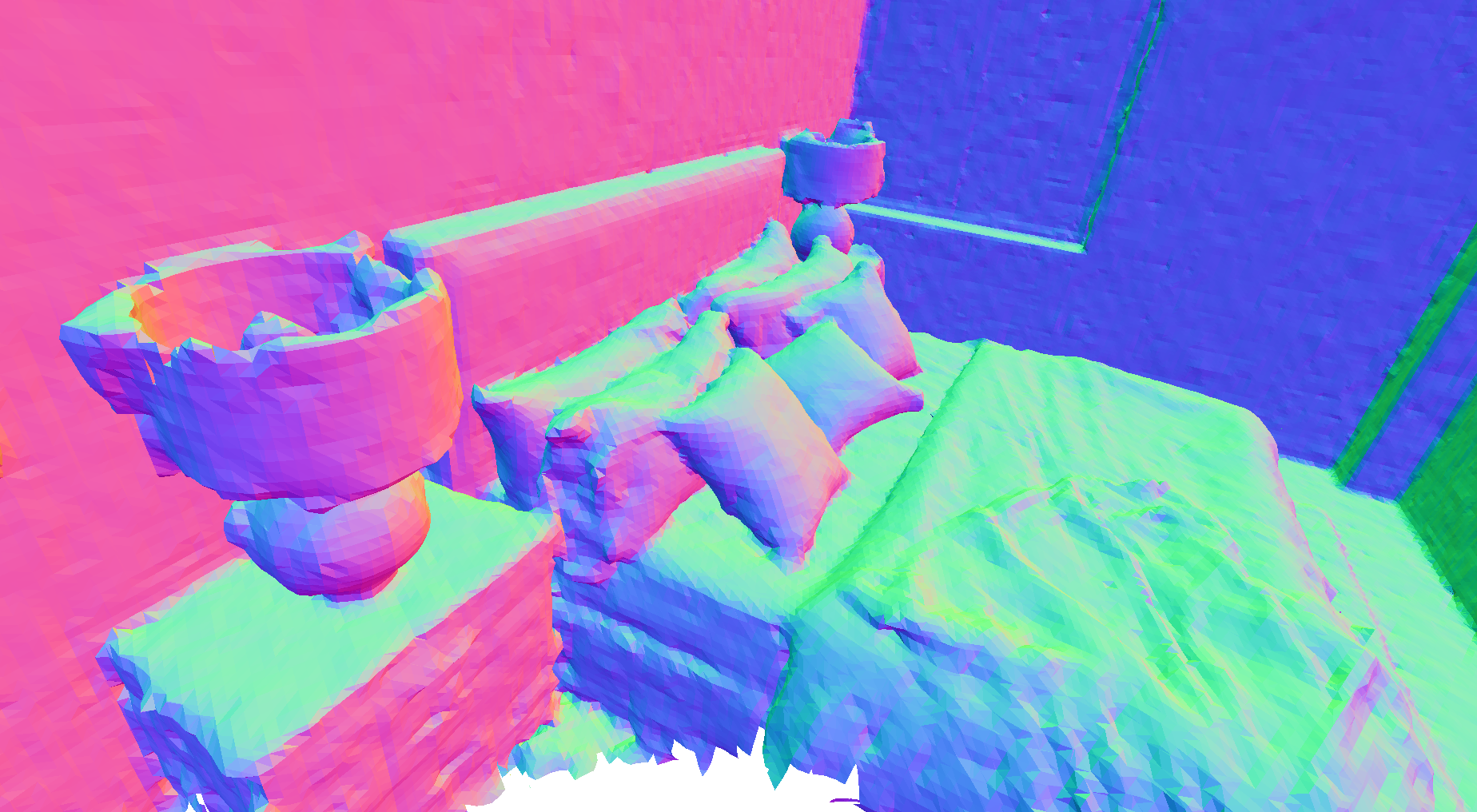}
    }
    \hspace{-11pt}
    \subfigure[Ground Truth]{
        \centering
        \includegraphics[width=0.162\textwidth]{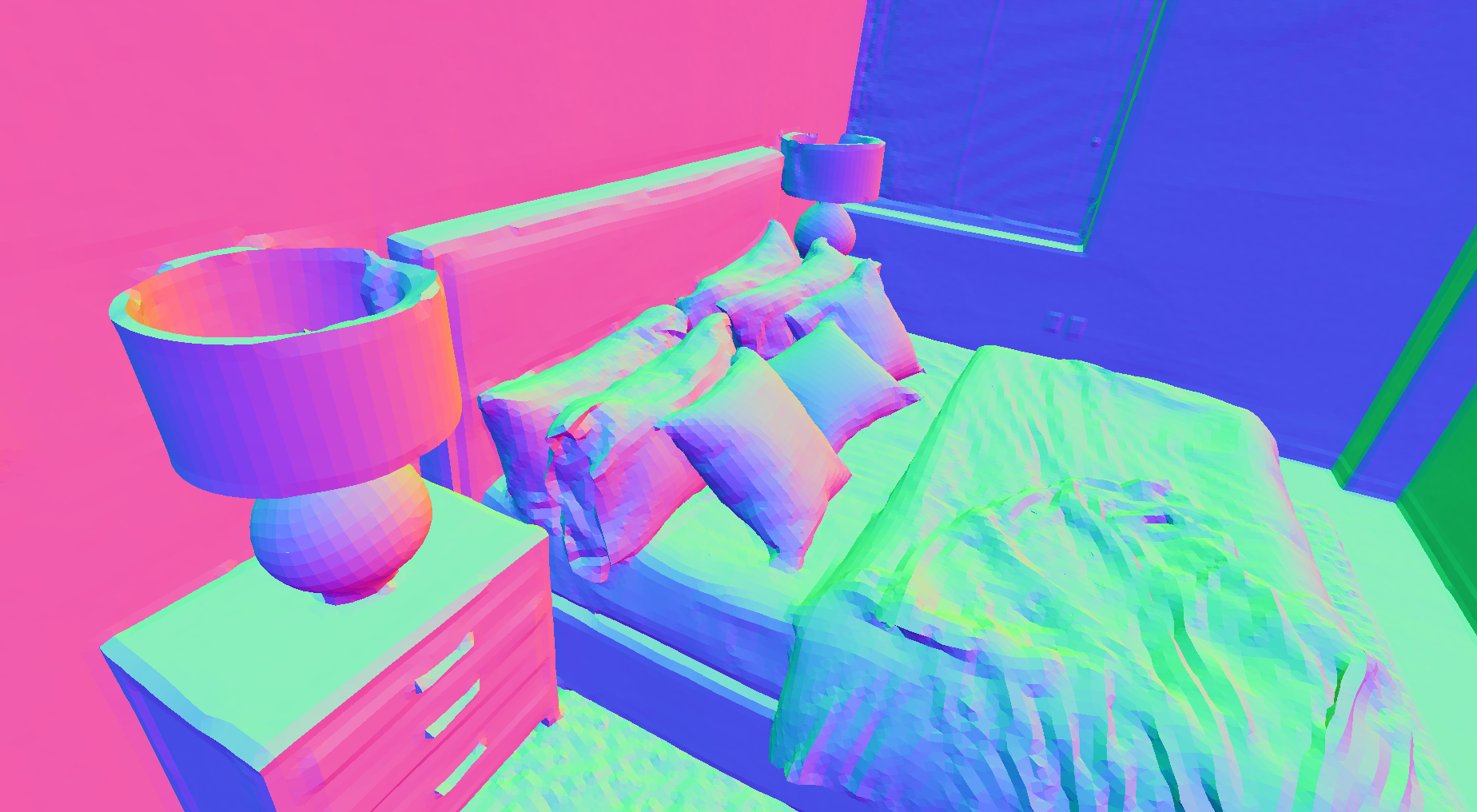}
    }
    \caption{
    Reconstructed mesh on the Replica dataset's office-2 and room-1, and colors indicate the direction of the surface normal. Our method can capture richer, more complete, and precise geometric details in small objects like tables, pillows, and quilts.}
    \label{fig:replica_qual}
    \vspace{-12pt}
    \end{figure*}

\subsection{Map Quality}\label{sec:map_quality}

\begin{figure*}[t]
     \subfigure[Voxblox]{
        \centering
        \includegraphics[width=0.18\textwidth]{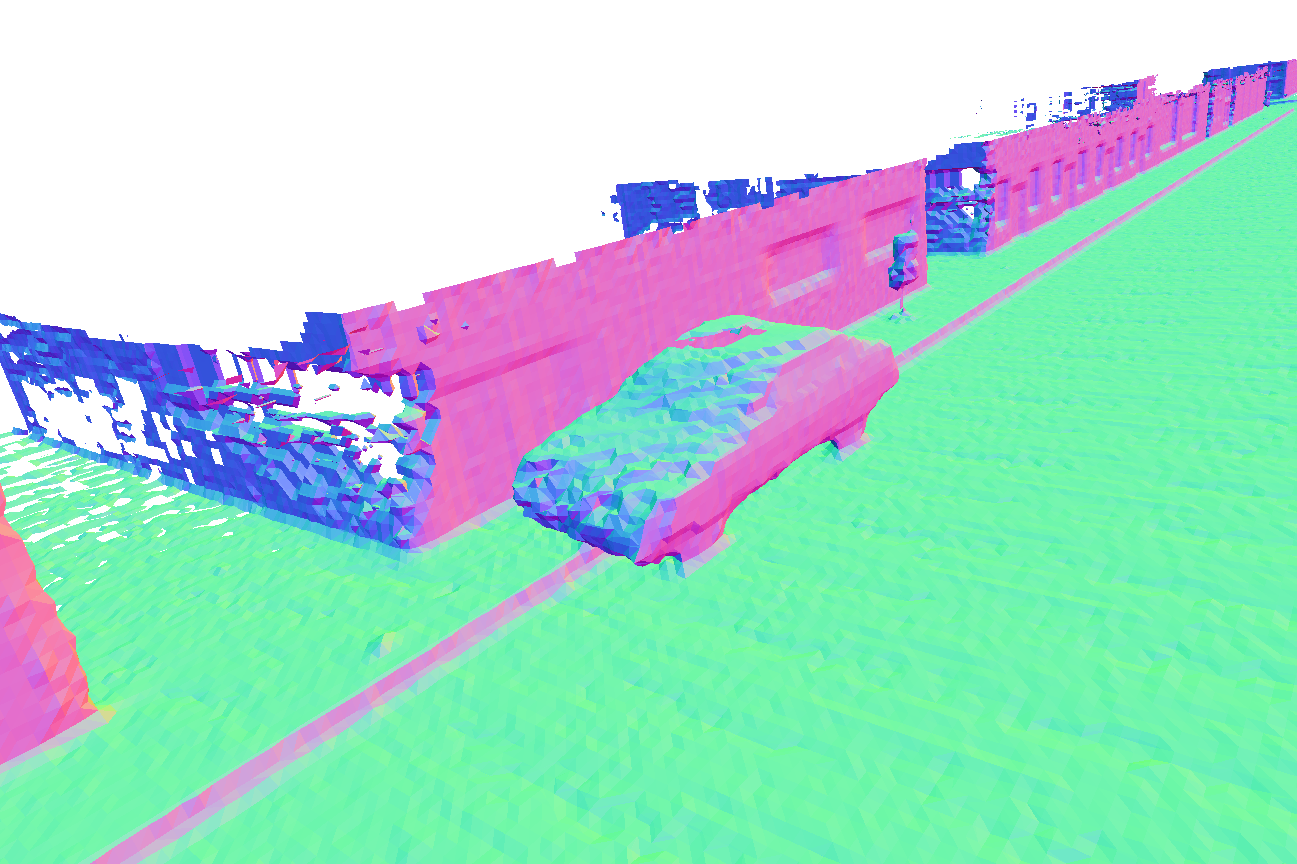}
     }
     \subfigure[VDB-Fusion]{
       \centering
       \includegraphics[width=0.18\textwidth]{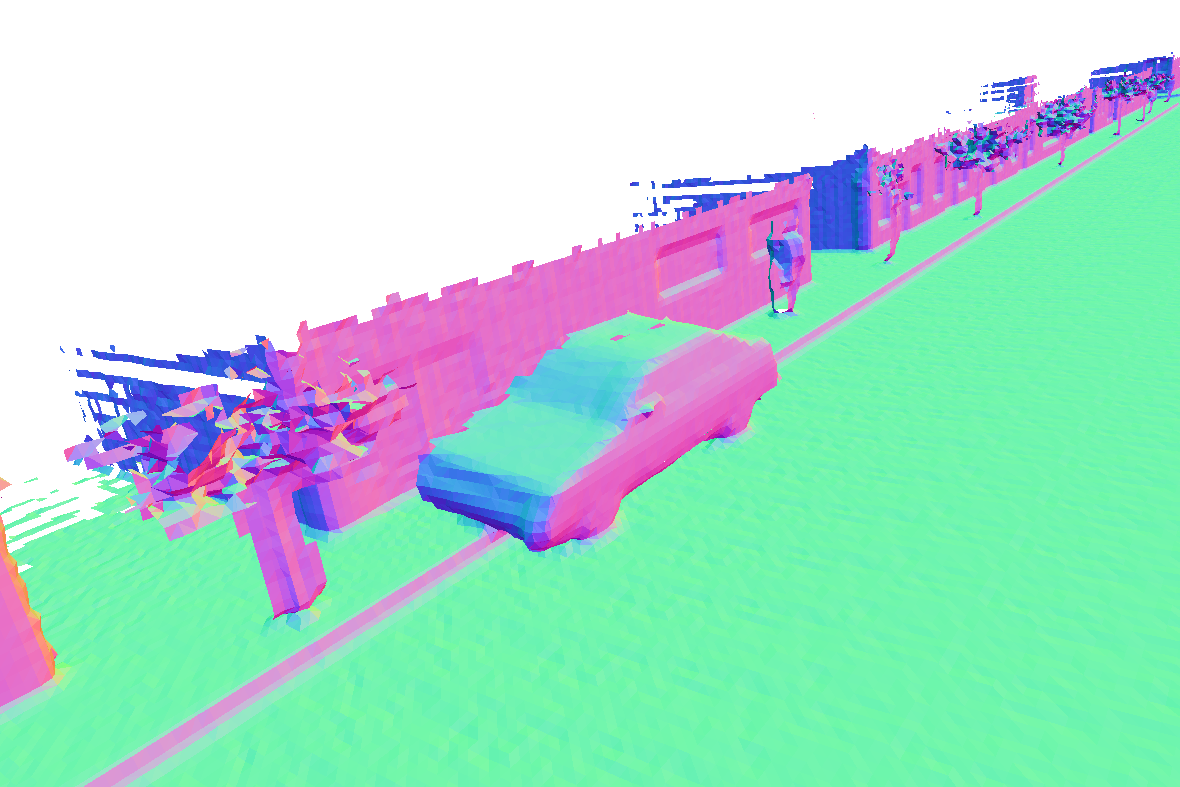}
    }
    \subfigure[SHINE-Mapping]{
       \centering
       \includegraphics[width=0.18\textwidth]{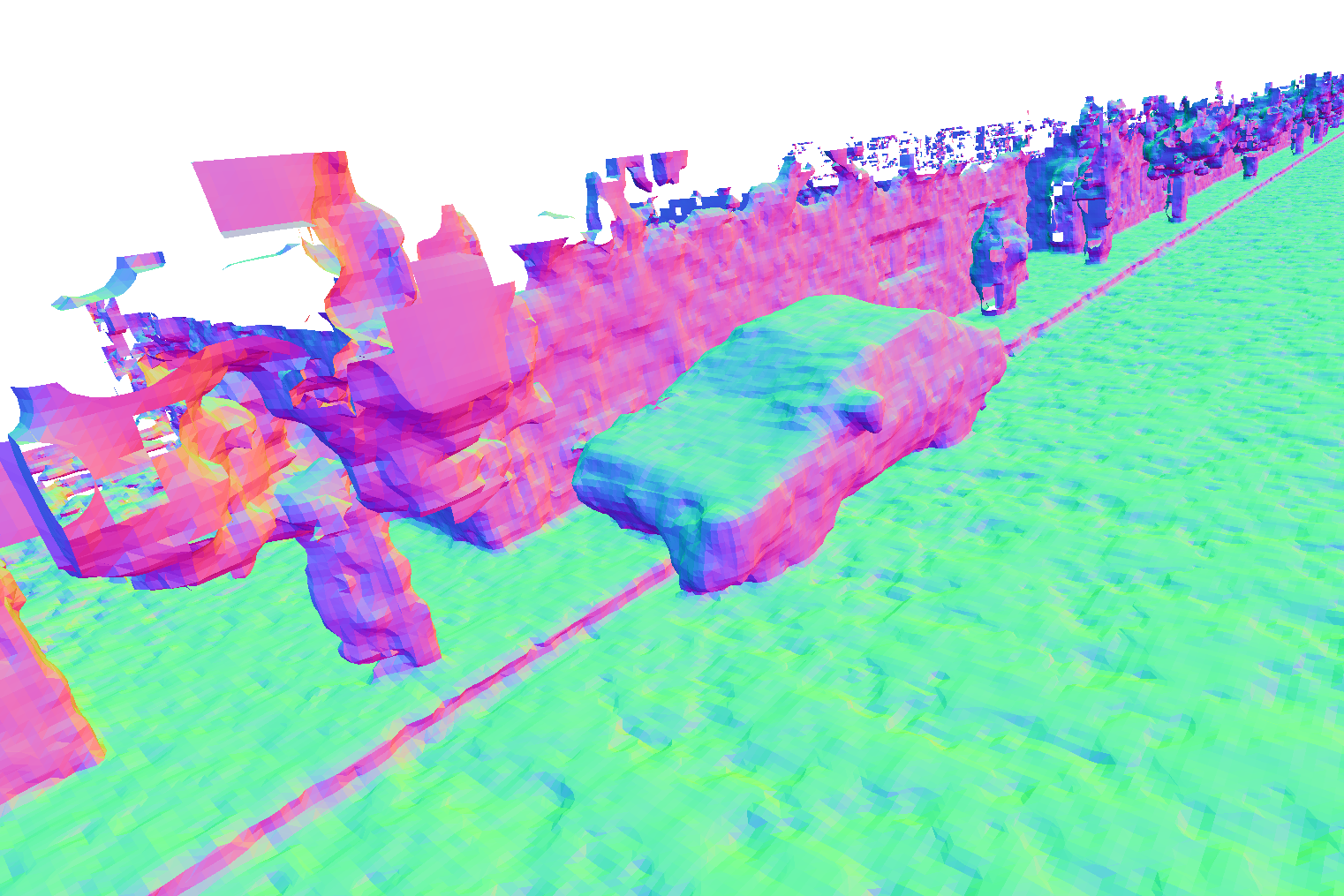}
     }
    \subfigure[Ours]{
       \centering
       \includegraphics[width=0.18\textwidth]{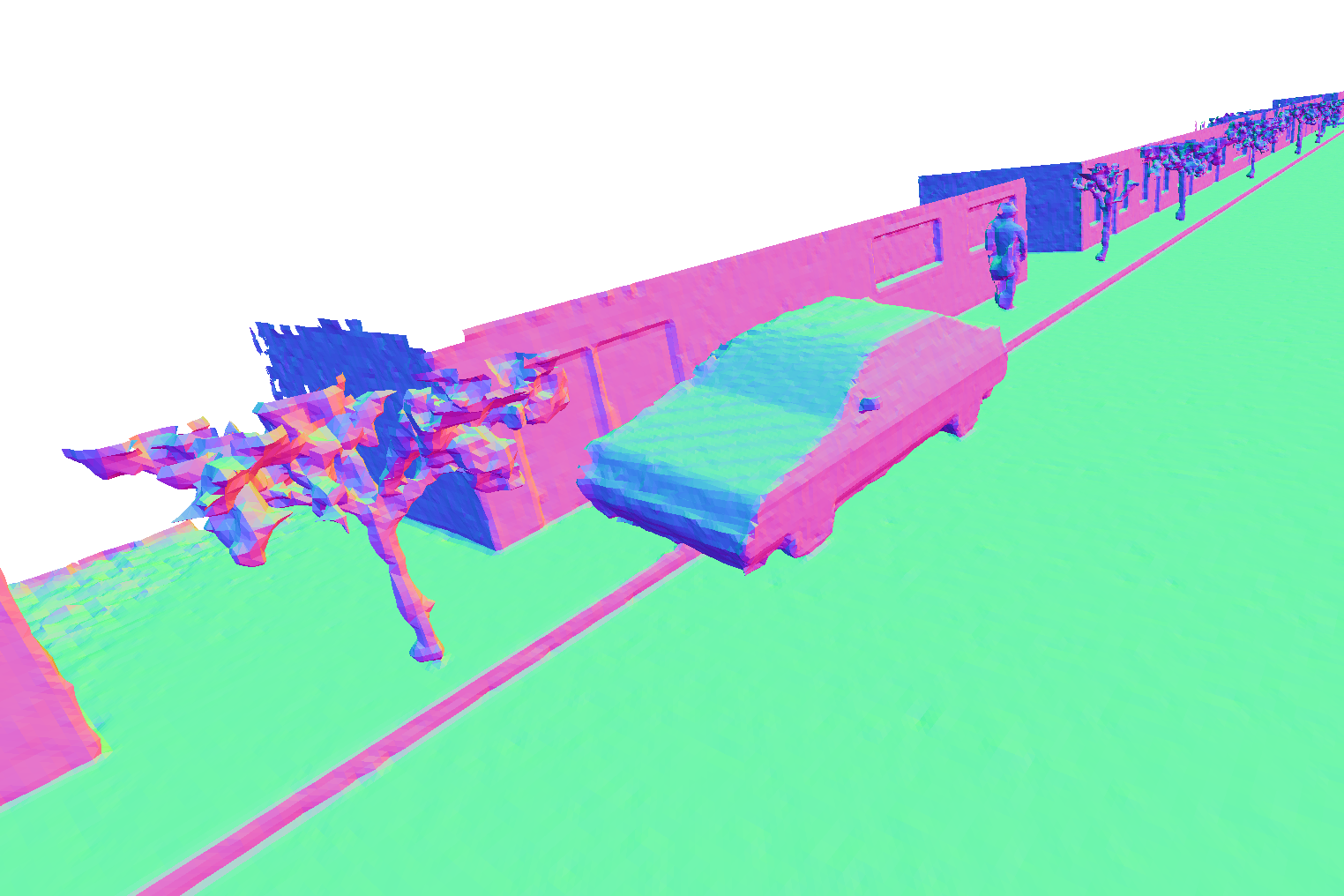}
    }
    \subfigure[Ours (w/o bundle)]{
       \centering
       \includegraphics[width=0.18\textwidth]{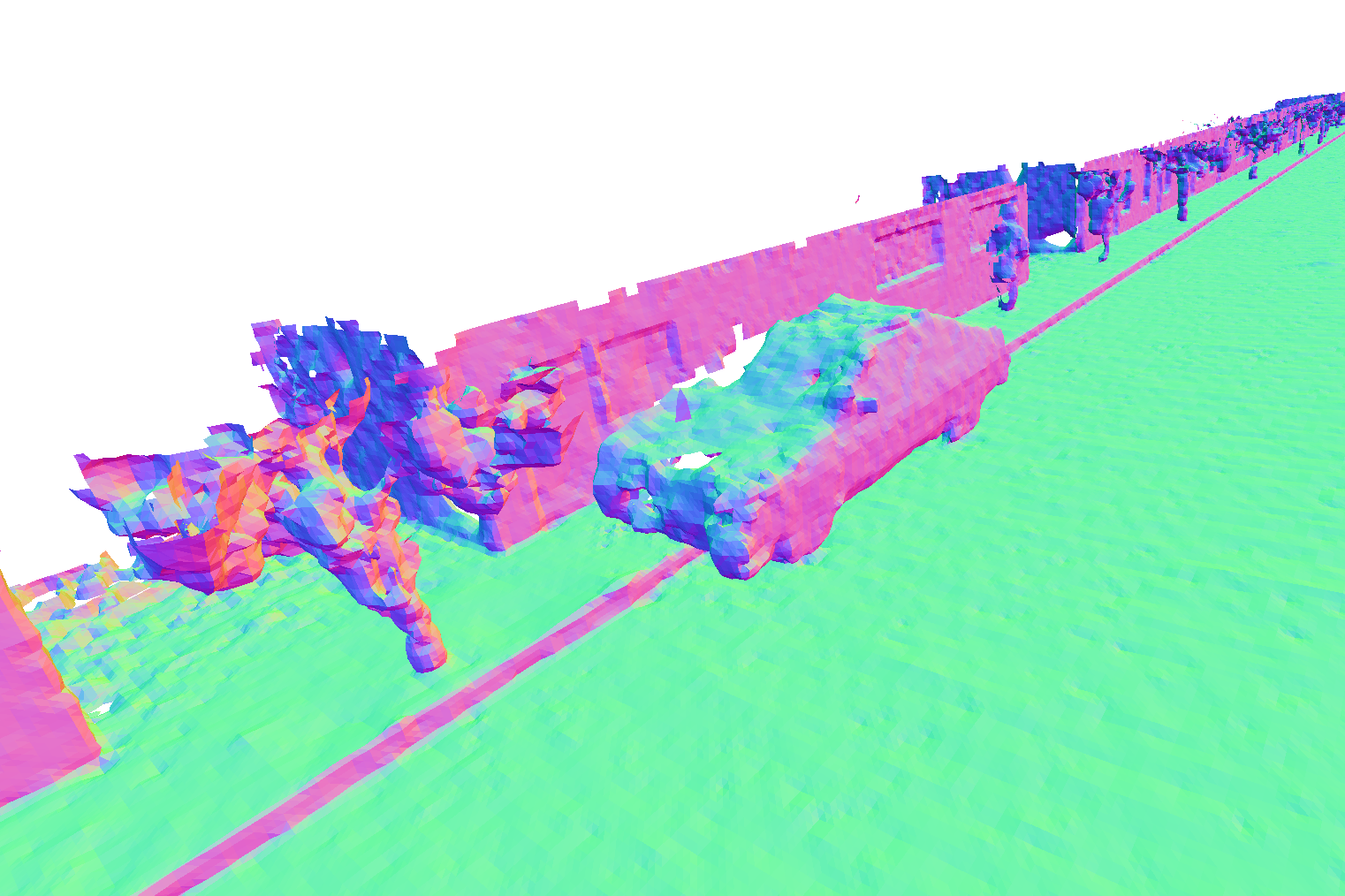}
    }
     
    \caption{Reconstructed mesh on the MaiCity dataset sequence 01, and colors indicate the surface normal direction. Our method outputs smoother and more complete mapping results, such as the trees, car, walker, walls, and the ground.}
    \label{fig:maicity_qual}
    \vspace{-8pt}
\end{figure*}

\subsubsection{Replica Dataset}

The Replica dataset \cite{straub2019replica} provides indoor simulation data generated by a camera mounted on a mobile manipulator. 
This experiment uses 8 of its scenarios for validation of indoor scene reconstruction. 
For small scenes, the leaf voxel size $s$ is set to $0.05$m, and $\sigma$ is set to $0.02$.
The quantitative reconstruction results are shown in Tab.~\ref{tab:replica_quan}. 
Regarding accuracy, VDBFusion performs best among all methods, with the lowest accuracy values for all scenes. 
Regarding completeness, SHINE performs best among all methods except for our proposed approach, with the lowest completeness values for most scenes.
The proposed approach strikes a good balance between accuracy and completeness, yielding the best overall performance in terms of C-L1 and F-Score.
The qualitative results from Fig.~\ref{fig:replica_qual} also indicate that explicit geometric representations show a sharper outline but lack completeness, and the learning-based implicit representations can speculate an unseen surface and output continuous mapping results at the expanse of details, especially for the pure neural-network-based method, iSDF. 

\begin{table}[t]
    \centering
    \caption{Quantitative reconstruction results on the MaiCIty dataset. And ablation studies for the proposed method's validation}
    \resizebox{0.48\textwidth}{!}{
      \begin{tabular}{clcccc}
          \toprule
          & \textbf{Methods}    & \textbf{Acc.$\downarrow$} & \textbf{Comp.$\downarrow$}                    & \textbf{C-L1$\downarrow$}  & \textbf{F-Score$\uparrow$}     \\ \midrule
          & Voxblox        & 2.554  & 2.875             & 2.715      & 97.261 \\
          & VDBFusion     & \textbf{2.091}  &  \underline{1.382}    & \underline{1.736}     & \underline{98.596} \\
          \midrule
          & SHINE  & 3.189 & {1.854}   & 2.521  &  95.472 \\
          & Ours           & \underline{2.167} & \textbf{0.958} & \textbf{1.562} & \textbf{98.923} \\  
          
          \midrule
          a) & random init.     & 2.982 & 1.118 & 2.050 & 96.082 \\  
          b) & w/o bundle           & 2.485 & 1.247 & 1.866 & 98.229 \\  
          c) & w/o free           & 2.217 & 0.963 & 1.590 & 98.723 \\   
          d) & w/o outside           & 2.213 & 0.988 & 1.600 & 98.835 \\  
          \midrule
          e) & Ours(online)    & 2.347 & 1.133 & 1.740 & 98.747 \\ 
          \bottomrule
          \end{tabular}
    }
    \label{tab:experiments_on_maicity}
    \vspace{-16pt}
\end{table}

\subsubsection{MaiCity Dataset}

We evaluate the urban scenario on sequence 01 of the MaiCity dataset\cite{vizzo2021poisson} that provides 100m noise-free 64-line lidar data and a ground truth model of a synthetic city. For large scenes, the leaf voxel size $s$ is set to $0.1$m, and $\sigma$ is set to $0.05$.
iSDF is not included as it is not scalable for large scenes. 
The quantitative and qualitative comparisons of different methods on the MaiCity dataset are shown in Tab.~\ref{tab:experiments_on_maicity} and Fig.~\ref{fig:maicity_qual}, respectively. 
It should be aware that SHINE-Mapping results come from its official incremental mapping with a regularization strategy, which differs from the results in its original paper\cite{zhong2022shine} using offline batch processing.
Our proposed method can produce smoother and more complete reconstruction results while retaining more fine-grained information. 
In particular, our method can generate flat surfaces on geometric structures, such as walls and floors, and also capture rich details on small objects, such as tree trunks and car rear-view mirrors. Voxblox and VDBFusion show decent accuracy but sacrifice completeness, like trees. However, SHINE-Mapping using a regularization-based method still faces forgetting issues in large-scale incremental mapping, resulting in inconsistent reconstruction results.

\subsubsection{Ablation Study}

To verify the effectiveness of the contributions, we conduct ablation studies and show the results in Tab.~\ref{tab:experiments_on_maicity}. 
As described in Sec.~\ref{sec:training}, \textbf{(a)} the random initialization of implicit features badly hinders the decoder from learning consistent signed distance fields. 
\textbf{(b)} Without bundle supervision (Sec.~\ref{sec:forgetting_problem}), both the reconstruction accuracy and completeness drop significantly, as shown in Fig.~\ref{fig:maicity_qual}\textbf{(e)}. 
The free \textbf{(c)} and outside \textbf{(d)} point supervisions (Sec.~\ref{sec:sampling}) slightly improve the mapping quality but help in mitigation of dynamic object influence (Sec.~\ref{sec_mapping_application}). 

\begin{figure}[htbp]
    \centering
    \includegraphics[width=0.46\textwidth]{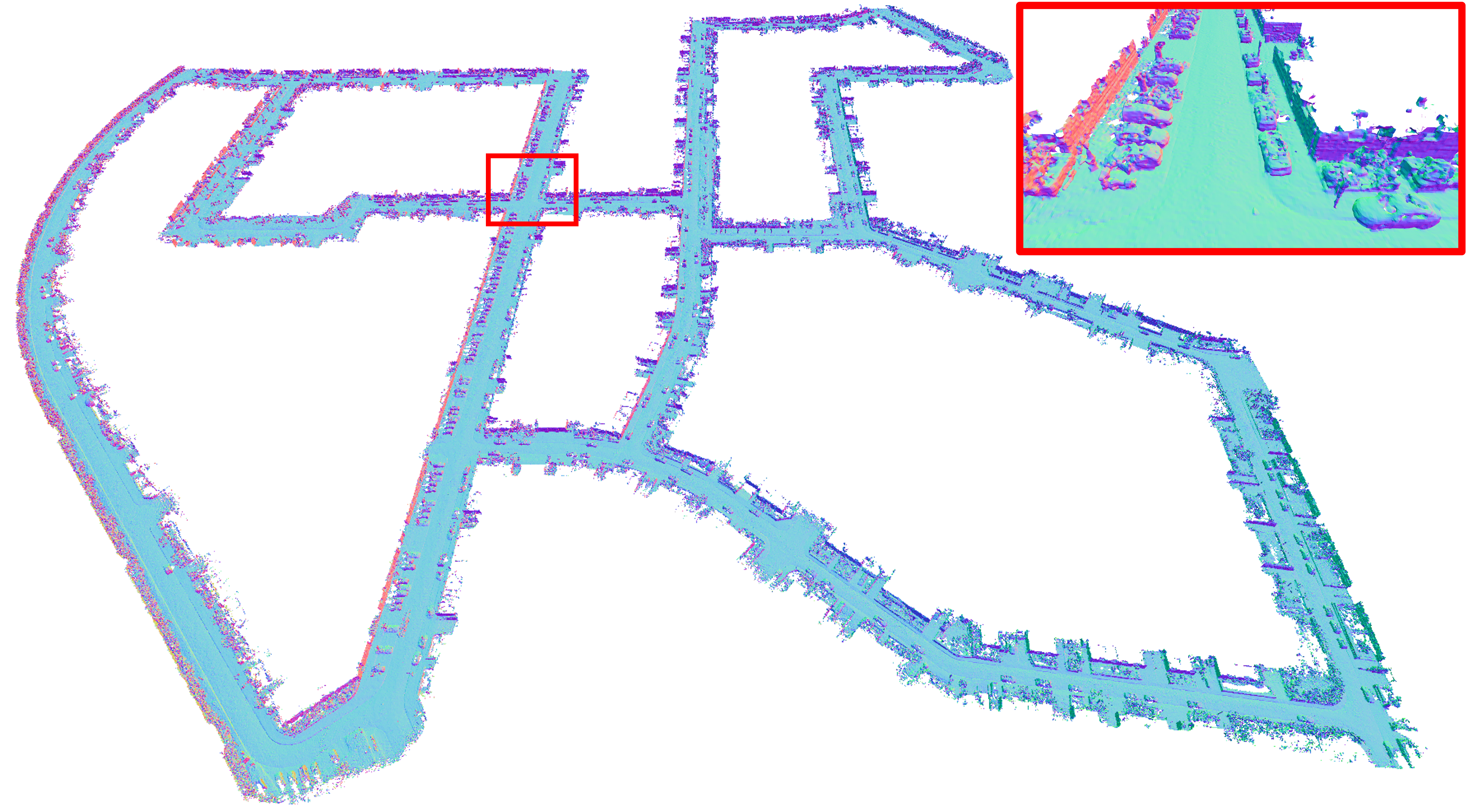}
    \caption{Reconstruction results on sequence 00 of the KITTI dataset, and the red box shows a local zoom-in view. Our proposed method can reconstruct high-quality, large-scale scenes using constant video memory.}
    \label{fig:kitti_qual}
    \vspace{-8pt}
\end{figure}

\subsection{Large-scale Incremental Mapping}\label{large_scale}

\begin{figure}[htbp]
    \centering
    \includegraphics[width=0.47\textwidth]{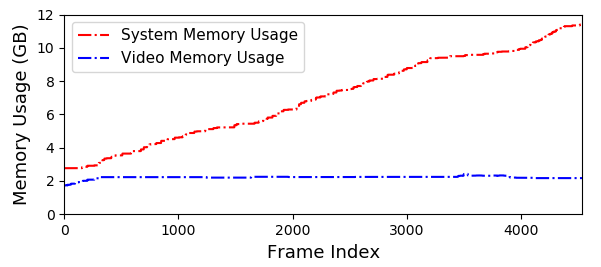}
    \caption{Memory usage of RIM on different devices. The video memory usage remains constant at the expense of the system memory usage.}
    \label{fig:kitti_mem}
\end{figure}

We demonstrate the unbounded reconstruction capability of RIM for large-scale scenes using the sequence 00 of the KITTI dataset\cite{geiger2012we}, which provides real-world LiDAR data with a distance of approximately 3.7 kilometers. 
RIM outputs a complete and smooth map as shown in Fig.~\ref{fig:kitti_qual}.
Fig.~\ref{fig:kitti_mem} shows the RIM's total software memory usage in different devices.
RIM can perform incremental mapping for extremely large outdoor scenes while avoiding exceeding video memory usage at the expanse of system memory usage thanks to the flexible map structure (Sec.~\ref{sec:structure}). 

\subsection{Runtime Analysis}

We evaluate the average integration time per frame on the MaiCity and KITTI datasets, as shown in Tab.~\ref{tab:runtime_comparison}.
SHINE-Mapping fails in reconstructing KITTI 00 due to the limited video memory of our experiment platform. 
RIM's default iteration number per frame is 50; when the iteration number is 10, RIM can process online and obtain competitive reconstruction quality, as shown in Tab.~\ref{tab:experiments_on_maicity}\textbf{(e)}. 
RIM's each module's spending times are shown in Tab.~\ref{tab:runtime}, wherein the inference process consists of encoding and forward processes.
The outlier removal module runs every second instead of every frame and takes an average of 8.56ms.
Our running efficiency and reconstruction quality are significantly better than SHINE-Mapping.
It is mainly contributed by the robot-centric mapping structure (Sec.~\ref{sec:structure}) and bundle supervision (Sec.~\ref{sec:forgetting_problem}).

\begin{table}[htbp]
    \centering
    \caption{Average integration time (ms) per frame on the MaiCity and KITTI dataset}
    \label{tab:runtime_comparison}
    {
    \begin{tabular}{cccc}
        \toprule
        \textbf{Datasets} & SHINE & Ours & Ours (online)\\
        \midrule
        MaiCity 01 & 1914.50 & 394.07 & 85.05\\
        KITTI 00 & - & 398.07 & {95.09}\\
        \bottomrule
    \end{tabular}
    }
    \vspace{-8pt}
\end{table}

\begin{table}[htbp]
    \centering
    \caption{Average time consumption (ms) of each module per frame}
    \label{tab:runtime}

    \resizebox{0.48\textwidth}{!}
    {
\begin{tabular}{cccccc}
    \toprule
    \multirow{2}*{Sliding} & \multirow{2}*{PreProcess} & \multicolumn{4}{c}{\textbf{Per Iteration}}\\
    \cmidrule{3-6}
    & & Sampling & Encoding & Forward & Backward\\
    \midrule
    6.06  & 0.76 & 3.88 & 1.40 & 0.10 & 2.40 \\
    \bottomrule
\end{tabular}
    }
\end{table}

\subsection{Mapping Application}\label{sec_mapping_application}
\begin{figure}[htbp]
    \centering

    \subfigure{
        \centering
        \includegraphics[width=0.225\textwidth]{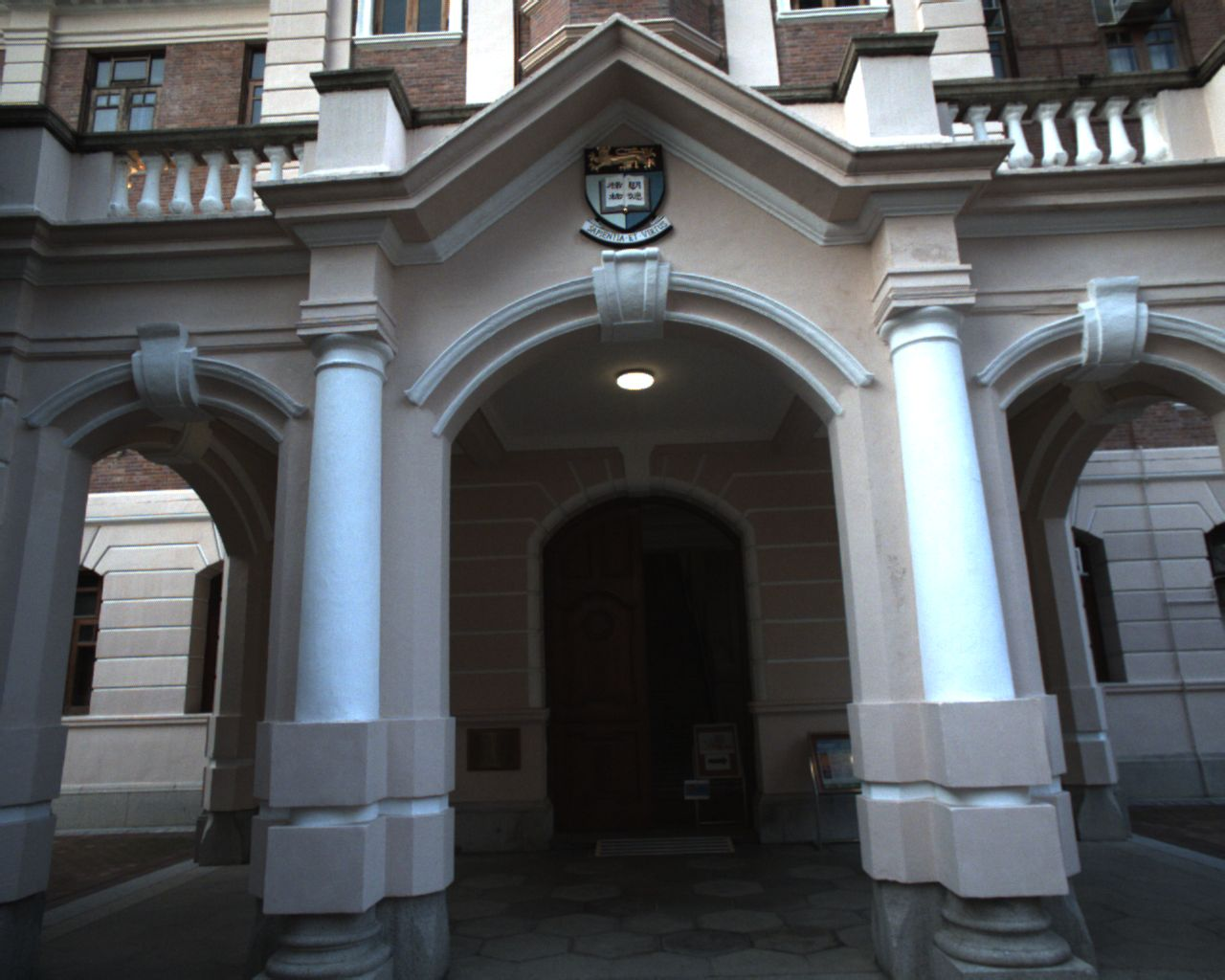}
    }
    \subfigure{
        \centering
        \includegraphics[width=0.225\textwidth]{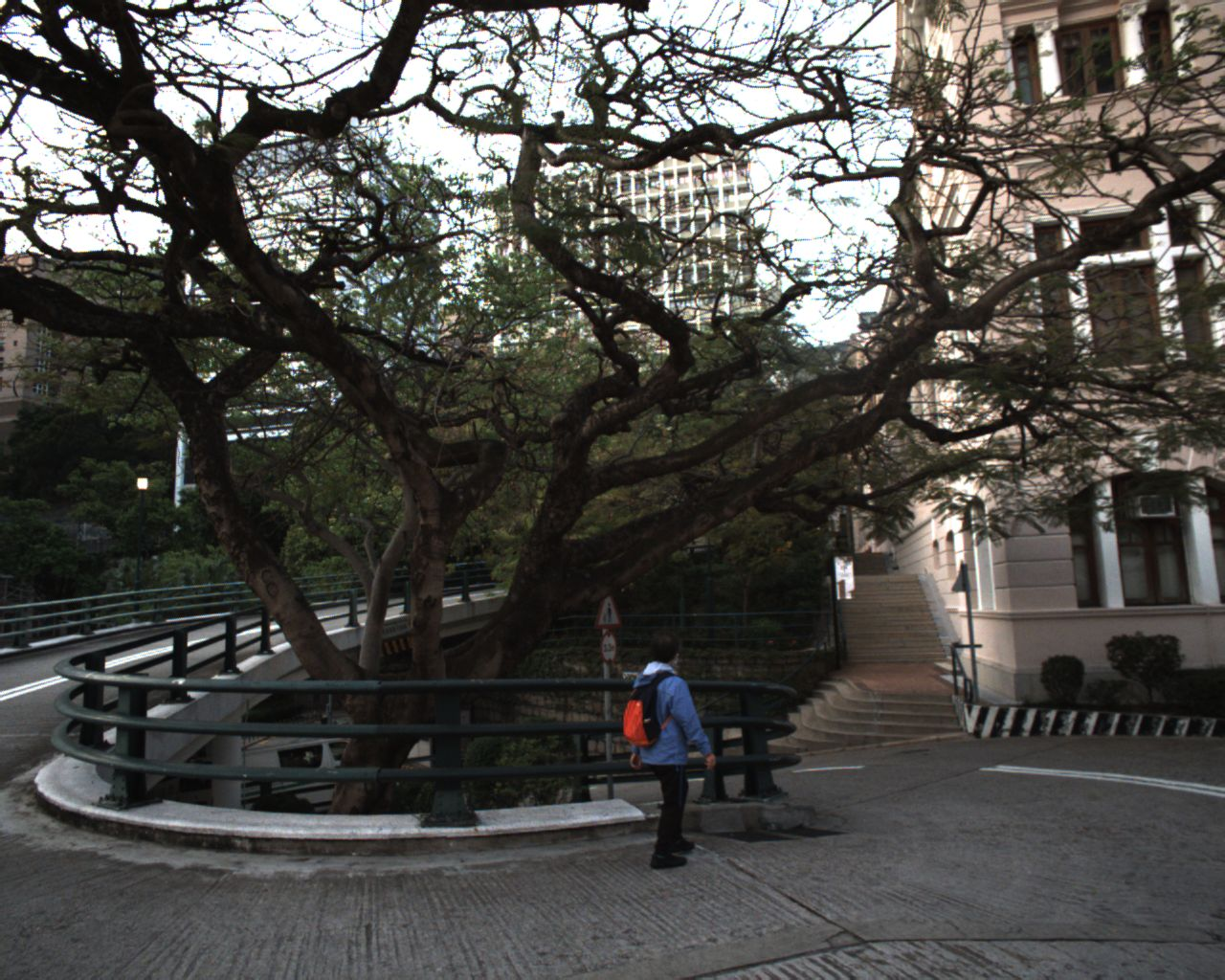}
    }
    \vspace{-14pt}
  
  \subfigure{
      \centering
      \includegraphics[width=0.225\textwidth]{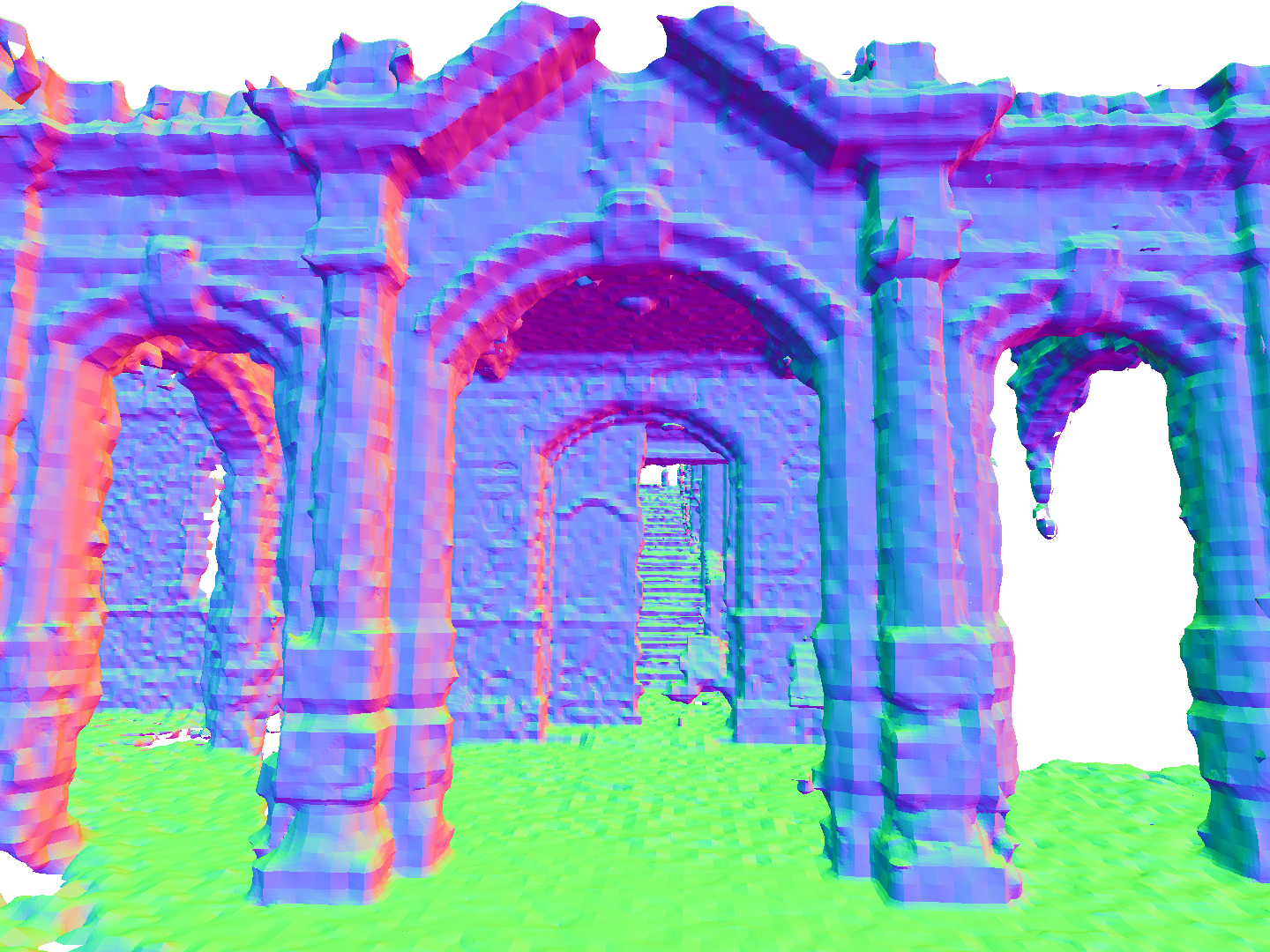}
  }
  \subfigure{
      \centering
      \includegraphics[width=0.225\textwidth]{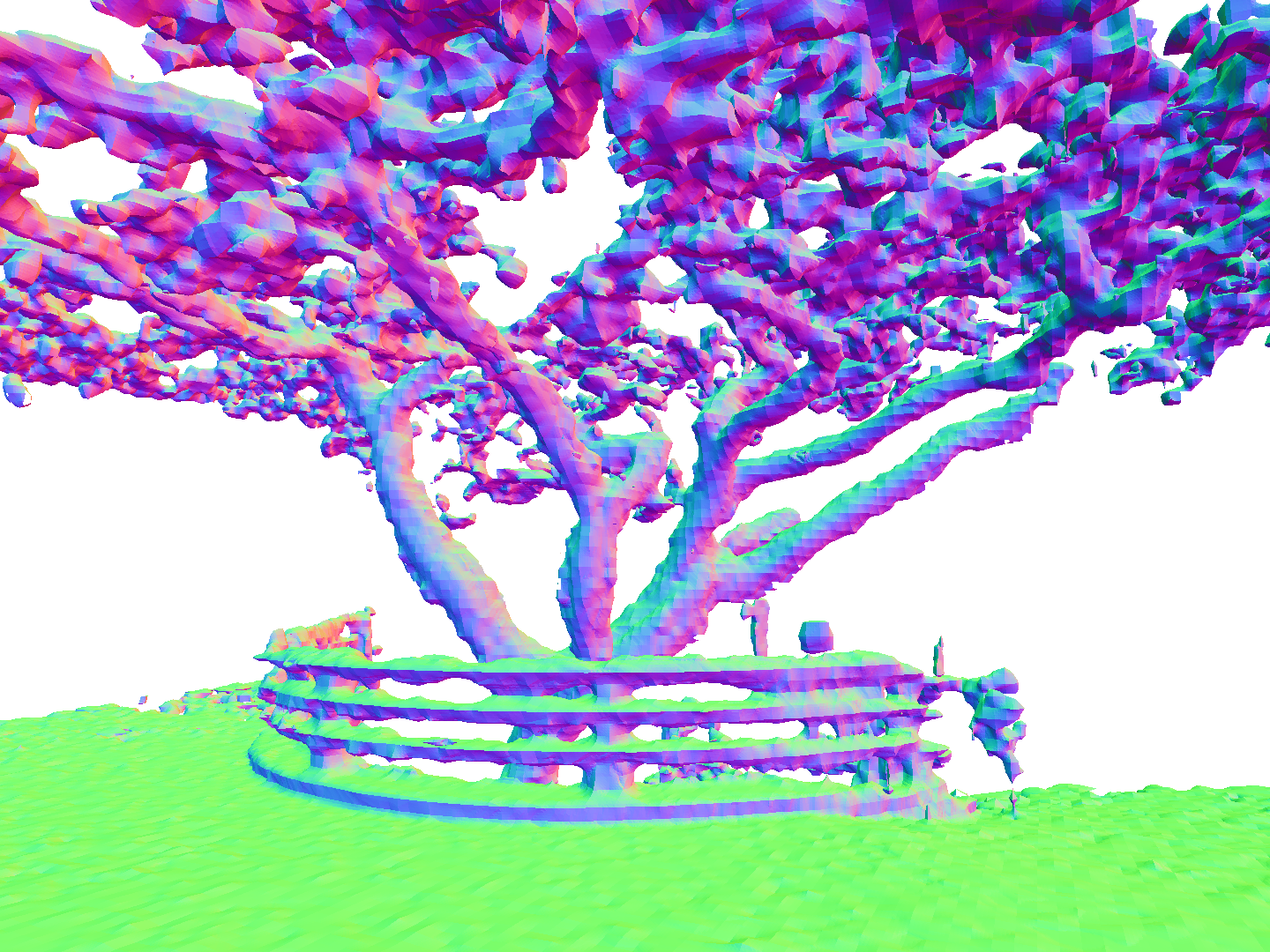}
  }
  \vspace{-14pt}

  \setcounter{subfigure}{0}
    \subfigure[HKU main building]{
      \centering
      \includegraphics[width=0.225\textwidth]{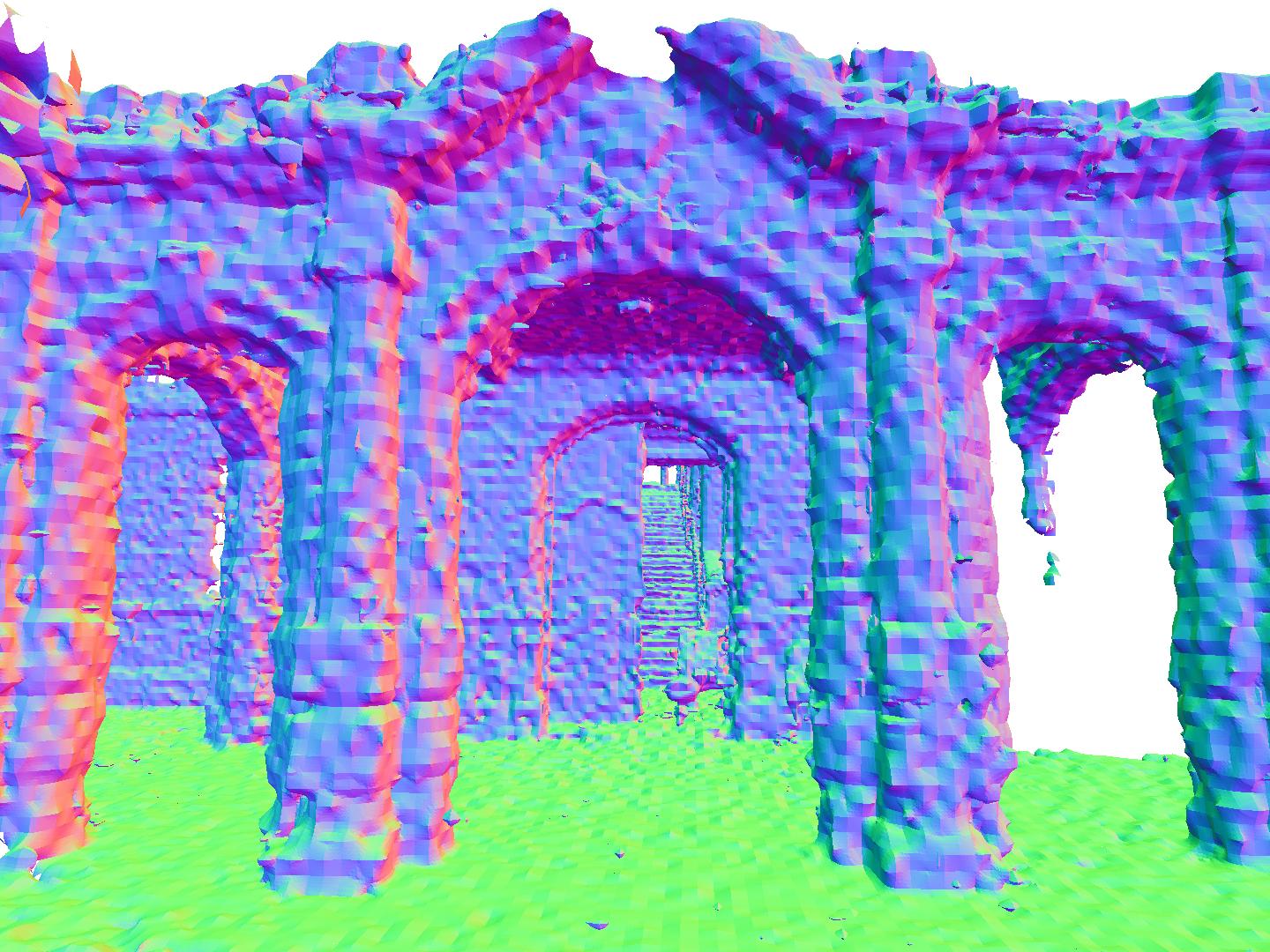}
  }
  \subfigure[Tree and walking person]{
      \centering
      \includegraphics[width=0.225\textwidth]{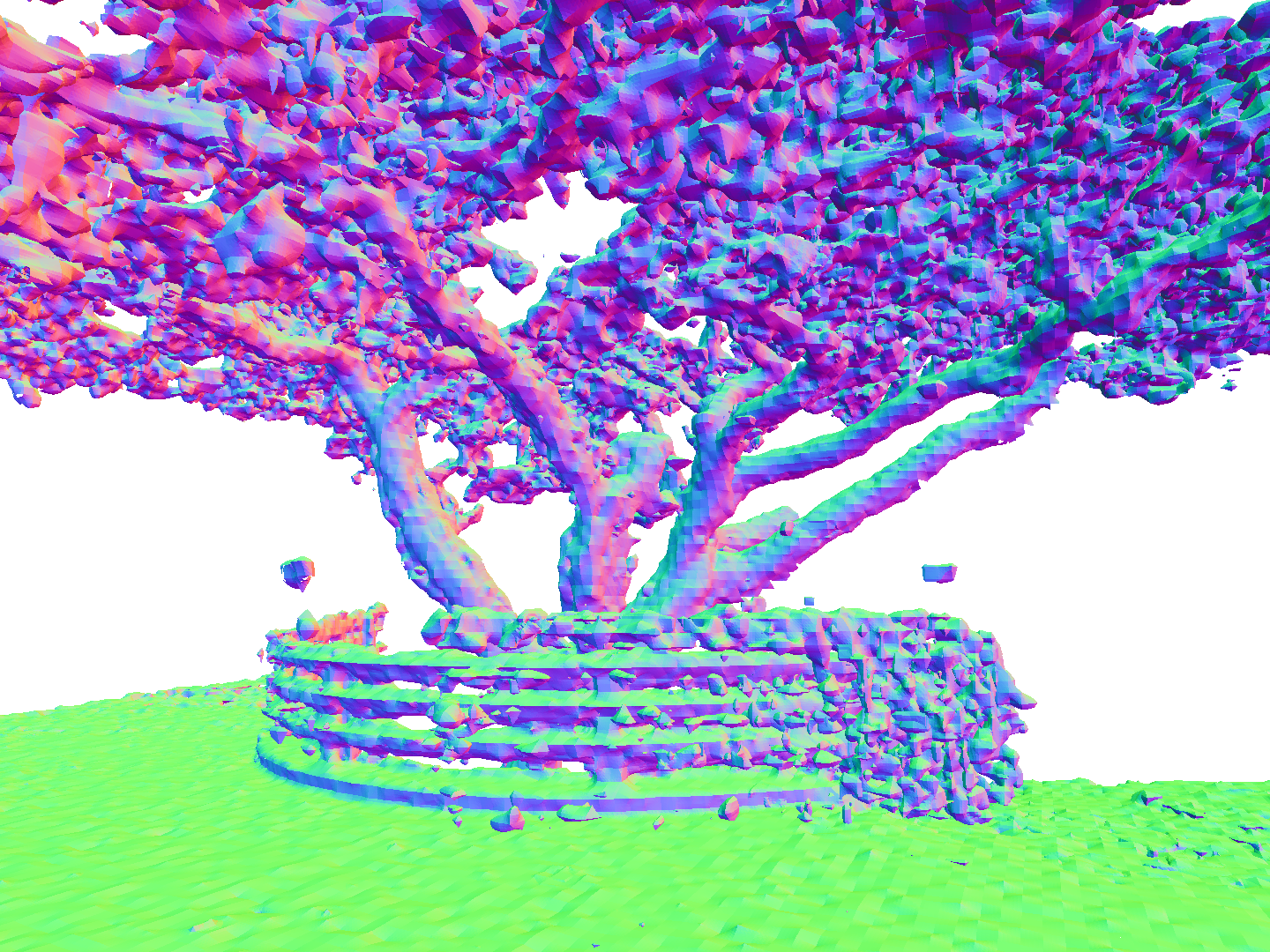}
  }

\caption{
    The reconstruction results on the HKU Main Building dataset, and the color represents the normal direction of the triangular surface. 
    The top row displays the original scenes in color, the middle row shows the RIM reconstruction results with the outlier removal module, and the bottom row displays results without the outlier removal module.
}
\label{fig:lio_rim_qual}
\end{figure}

This section presents a real-time incremental mapping application employing RIM with lidar inertial odometry\cite{xu2022fast}. It is assessed using the HKU Main Building dataset\cite{lin2022r}, which provides real-world lidar data sourced from a solid-state lidar RIM can mitigate the impact of noisy input with the help of outlier removal (Sec.~\ref{sec:forgetting_problem}), producing commendable, detailed reconstruction results as depicted in Fig.~\ref{fig:lio_rim_qual}. Noisy outliers are eliminated within the main building scene to yield a refined reconstruction. In situations involving trees and a traverse-walking pedestrian, our approach significantly reduces the influence of dynamic objects.

\section{Conclusion}

This paper introduced an efficient, scalable, robot-centric implicit mapping using range sensors. Key to our system is the subtle leverage of the robot-centric local map, which enhances both implicit model training efficiency and overall mapping quality while addressing the typical catastrophic forgetting phenomenon encountered in continual learning. We employ a flexible global map to preserve a reusable map. Our method, evaluated on various datasets, has proven to excel in terms of reconstruction quality, efficiency, and adaptability.
Our method is tailored to mapping and depends on the precision of poses. 
Our future endeavors will include the incorporation of pose refinement during training and deploying our method into robotics tasks. 



{
\bibliographystyle{IEEEtran}
\balance
\bibliography{reference}

\begin{thebibliography}{10}
\providecommand{\url}[1]{#1}
\csname url@rmstyle\endcsname
\providecommand{\newblock}{\relax}
\providecommand{\bibinfo}[2]{#2}
\providecommand\BIBentrySTDinterwordspacing{\spaceskip=0pt\relax}
\providecommand\BIBentryALTinterwordstretchfactor{4}
\providecommand\BIBentryALTinterwordspacing{\spaceskip=\fontdimen2\font plus
\BIBentryALTinterwordstretchfactor\fontdimen3\font minus
  \fontdimen4\font\relax}
\providecommand\BIBforeignlanguage[2]{{%
\expandafter\ifx\csname l@#1\endcsname\relax
\typeout{** WARNING: IEEEtran.bst: No hyphenation pattern has been}%
\typeout{** loaded for the language `#1'. Using the pattern for}%
\typeout{** the default language instead.}%
\else
\language=\csname l@#1\endcsname
\fi
#2}}

\bibitem{hornung2013octomap}
A.~Hornung, K.~M. Wurm, M.~Bennewitz, C.~Stachniss, and W.~Burgard, ``Octomap:
  An efficient probabilistic 3d mapping framework based on octrees,''
  \emph{Autonomous robots}, vol.~34, pp. 189--206, 2013.

\bibitem{museth2013vdb}
K.~Museth, ``Vdb: High-resolution sparse volumes with dynamic topology,''
  \emph{ACM Transactions on Graphics (TOG)}, vol.~32, no.~3, pp. 1--22, 2013.

\bibitem{niessner2013real}
M.~Nie{\ss}ner, M.~Zollh{\"o}fer, S.~Izadi, and M.~Stamminger, ``Real-time 3d
  reconstruction at scale using voxel hashing,'' \emph{ACM Transactions on
  Graphics (ToG)}, vol.~32, no.~6, pp. 1--11, 2013.

\bibitem{oleynikova2017voxblox}
H.~Oleynikova, Z.~Taylor, M.~Fehr, R.~Siegwart, and J.~Nieto, ``Voxblox:
  Incremental 3d euclidean signed distance fields for on-board mav planning,''
  in \emph{2017 IEEE/RSJ International Conference on Intelligent Robots and
  Systems (IROS)}.\hskip 1em plus 0.5em minus 0.4em\relax IEEE, 2017, pp.
  1366--1373.

\bibitem{vizzo2022vdbfusion}
I.~Vizzo, T.~Guadagnino, J.~Behley, and C.~Stachniss, ``Vdbfusion: Flexible and
  efficient tsdf integration of range sensor data,'' \emph{Sensors}, vol.~22,
  no.~3, p. 1296, 2022.

\bibitem{mescheder2019occupancy}
L.~Mescheder, M.~Oechsle, M.~Niemeyer, S.~Nowozin, and A.~Geiger, ``Occupancy
  networks: Learning 3d reconstruction in function space,'' in
  \emph{Proceedings of the IEEE/CVF Conference on Computer Vision and Pattern
  Recognition}, 2019, pp. 4460--4470.

\bibitem{mildenhall2021nerf}
B.~Mildenhall, P.~P. Srinivasan, M.~Tancik, J.~T. Barron, R.~Ramamoorthi, and
  R.~Ng, ``Nerf: Representing scenes as neural radiance fields for view
  synthesis,'' \emph{Communications of the ACM}, vol.~65, no.~1, pp. 99--106,
  2021.

\bibitem{niemeyer2020differentiable}
M.~Niemeyer, L.~Mescheder, M.~Oechsle, and A.~Geiger, ``Differentiable
  volumetric rendering: Learning implicit 3d representations without 3d
  supervision,'' in \emph{Proceedings of the IEEE/CVF Conference on Computer
  Vision and Pattern Recognition}, 2020, pp. 3504--3515.

\bibitem{park2019deepsdf}
J.~J. Park, P.~Florence, J.~Straub, R.~Newcombe, and S.~Lovegrove, ``Deepsdf:
  Learning continuous signed distance functions for shape representation,'' in
  \emph{Proceedings of the IEEE/CVF Conference on Computer Vision and Pattern
  Recognition}, 2019, pp. 165--174.

\bibitem{zhang2020nerf++}
K.~Zhang, G.~Riegler, N.~Snavely, and V.~Koltun, ``Nerf++: Analyzing and
  improving neural radiance fields,'' \emph{arXiv preprint arXiv:2010.07492},
  2020.

\bibitem{ortiz2022isdf}
J.~Ortiz, A.~Clegg, J.~Dong, E.~Sucar, D.~Novotny, M.~Zollhoefer, and
  M.~Mukadam, ``isdf: Real-time neural signed distance fields for robot
  perception,'' \emph{arXiv preprint arXiv:2204.02296}, 2022.

\bibitem{liu2020neural}
L.~Liu, J.~Gu, K.~Zaw~Lin, T.-S. Chua, and C.~Theobalt, ``Neural sparse voxel
  fields,'' \emph{Advances in Neural Information Processing Systems}, vol.~33,
  pp. 15\,651--15\,663, 2020.

\bibitem{takikawa2021neural}
T.~Takikawa, J.~Litalien, K.~Yin, K.~Kreis, C.~Loop, D.~Nowrouzezahrai,
  A.~Jacobson, M.~McGuire, and S.~Fidler, ``Neural geometric level of detail:
  Real-time rendering with implicit 3d shapes,'' in \emph{Proceedings of the
  IEEE/CVF Conference on Computer Vision and Pattern Recognition}, 2021, pp.
  11\,358--11\,367.

\bibitem{muller2022instant}
T.~M{\"u}ller, A.~Evans, C.~Schied, and A.~Keller, ``Instant neural graphics
  primitives with a multiresolution hash encoding,'' \emph{arXiv preprint
  arXiv:2201.05989}, 2022.

\bibitem{zhong2022shine}
X.~Zhong, Y.~Pan, J.~Behley, and C.~Stachniss, ``Shine-mapping: Large-scale 3d
  mapping using sparse hierarchical implicit neural representations,''
  \emph{arXiv preprint arXiv:2210.02299}, 2022.

\bibitem{yan2023active}
D.~Yan, J.~Liu, F.~Quan, H.~Chen, and M.~Fu, ``Active implicit object
  reconstruction using uncertainty-guided next-best-view optimization,''
  \emph{IEEE Robotics and Automation Letters}, pp. 1--8, 2023.

\bibitem{kaushik2021understanding}
P.~Kaushik, A.~Gain, A.~Kortylewski, and A.~Yuille, ``Understanding
  catastrophic forgetting and remembering in continual learning with optimal
  relevance mapping,'' \emph{arXiv preprint arXiv:2102.11343}, 2021.

\bibitem{sucar2021imap}
E.~Sucar, S.~Liu, J.~Ortiz, and A.~J. Davison, ``imap: Implicit mapping and
  positioning in real-time,'' in \emph{Proceedings of the IEEE/CVF
  International Conference on Computer Vision}, 2021, pp. 6229--6238.

\bibitem{zhu2022nice}
Z.~Zhu, S.~Peng, V.~Larsson, W.~Xu, H.~Bao, Z.~Cui, M.~R. Oswald, and
  M.~Pollefeys, ``Nice-slam: Neural implicit scalable encoding for slam,'' in
  \emph{Proceedings of the IEEE/CVF Conference on Computer Vision and Pattern
  Recognition}, 2022, pp. 12\,786--12\,796.

\bibitem{liu2022rgb}
J.~Liu, X.~Li, Y.~Liu, and H.~Chen, ``Rgb-d inertial odometry for a
  resource-restricted robot in dynamic environments,'' \emph{IEEE Robotics and
  Automation Letters}, vol.~7, no.~4, pp. 9573--9580, 2022.

\bibitem{newcombe2011kinectfusion}
R.~A. Newcombe, S.~Izadi, O.~Hilliges, D.~Molyneaux, D.~Kim, A.~J. Davison,
  P.~Kohi, J.~Shotton, S.~Hodges, and A.~Fitzgibbon, ``Kinectfusion: Real-time
  dense surface mapping and tracking,'' in \emph{2011 10th IEEE International
  Symposium on Mixed and Augmented Reality}.\hskip 1em plus 0.5em minus
  0.4em\relax Ieee, 2011, pp. 127--136.

\bibitem{deng2022depth}
K.~Deng, A.~Liu, J.-Y. Zhu, and D.~Ramanan, ``Depth-supervised nerf: Fewer
  views and faster training for free,'' in \emph{Proceedings of the IEEE/CVF
  Conference on Computer Vision and Pattern Recognition}, 2022, pp.
  12\,882--12\,891.

\bibitem{lorensen1987marching}
W.~E. Lorensen and H.~E. Cline, ``Marching cubes: A high resolution 3d surface
  construction algorithm,'' \emph{ACM Siggraph Computer Graphics}, vol.~21,
  no.~4, pp. 163--169, 1987.

\bibitem{straub2019replica}
J.~Straub, T.~Whelan, L.~Ma, Y.~Chen, E.~Wijmans, S.~Green, J.~J. Engel,
  R.~Mur-Artal, C.~Ren, S.~Verma, \emph{et~al.}, ``The replica dataset: A
  digital replica of indoor spaces,'' \emph{arXiv preprint arXiv:1906.05797},
  2019.

\bibitem{vizzo2021poisson}
I.~Vizzo, X.~Chen, N.~Chebrolu, J.~Behley, and C.~Stachniss, ``Poisson surface
  reconstruction for lidar odometry and mapping,'' in \emph{2021 IEEE
  International Conference on Robotics and Automation (ICRA)}.\hskip 1em plus
  0.5em minus 0.4em\relax IEEE, 2021, pp. 5624--5630.

\bibitem{geiger2012we}
A.~Geiger, P.~Lenz, and R.~Urtasun, ``Are we ready for autonomous driving? the
  kitti vision benchmark suite,'' in \emph{2012 IEEE Conference on Computer
  Vision and Pattern Recognition}.\hskip 1em plus 0.5em minus 0.4em\relax IEEE,
  2012, pp. 3354--3361.

\bibitem{xu2022fast}
W.~Xu, Y.~Cai, D.~He, J.~Lin, and F.~Zhang, ``Fast-lio2: Fast direct
  lidar-inertial odometry,'' \emph{IEEE Transactions on Robotics}, vol.~38,
  no.~4, pp. 2053--2073, 2022.

\bibitem{lin2022r}
J.~Lin and F.~Zhang, ``R 3 live: A robust, real-time, rgb-colored,
  lidar-inertial-visual tightly-coupled state estimation and mapping package,''
  in \emph{2022 International Conference on Robotics and Automation
  (ICRA)}.\hskip 1em plus 0.5em minus 0.4em\relax IEEE, 2022, pp.
  10\,672--10\,678.

\end{thebibliography}
}

\end{document}